\documentclass[10pt,3p]{elsarticle}

\usepackage{hyperref}

\usepackage{times}
\usepackage{epsfig}
\usepackage{algorithm}
\usepackage{graphicx}
\usepackage{amsmath,amssymb} %
\usepackage{color}

\usepackage{natbib}
\usepackage{xspace,eucal,dsfont,url}
\usepackage{wrapfig,lipsum,booktabs}
\usepackage{multirow}

\def\onedot{.\xspace}
\def\eg{\emph{e.g}\onedot}

\def\etc{\emph{etc}\onedot}

\def\etal{\emph{et al}\onedot}
\def\half{{\tfrac{1}{2} }}

\def\argmin{\operatorname*{argmin\,}}
\def\T{{\!\top}}

\def\Real{\mathbb{R}}

\newcommand{\ssvm}{{\sc ssvm}\xspace}
\newcommand{\svm}{{\sc svm}\xspace}

\newcommand{\crf}{{\sc crf}\xspace}
\newcommand{\crfs}{{\sc crf}s\xspace}

\def\bw{{\bf w}}
\def\bx{{\bf x}}
\def\by{{\bf y}}

\def\bh{{ \Phi }}
\let\x\bx
\let\y\by
\let\w\bw

\def\z{{\bf y}'}

\def\bxi{{\boldsymbol \xi}}

\def\calY{{\cal Y}}

\def\loss{{\it \Delta}}

\def\X{ {\bf X } }
\def\Y{ {\bf Y } }

\newcommand{\st}{{{\rm s.t.}\!:}\xspace}
\newcommand{\fnorm}[2][2]{\ensuremath{ \left\| #2 \right\|_{ \mathrm{#1} } } }

\begin{document}

\begin{frontmatter}

\title{CRF Learning with CNN Features for Image Segmentation}

\author{Fayao Liu, Guosheng Lin, Chunhua Shen}
\address{School of Computer Science, The University of Adelaide, Australia}

\begin{abstract}
Conditional Random Rields (\crf) have been widely applied in image segmentations. While most studies rely on hand-crafted features,
we here propose to exploit a pre-trained large convolutional neural network (CNN) to generate deep features for \crf learning. The deep CNN is trained on the ImageNet dataset and transferred to image segmentations here for constructing potentials of superpixels.
Then the \crf parameters are learnt using a structured support vector machine (\ssvm).
To fully exploit context information in inference, we construct spatially related co-occurrence pairwise potentials and incorporate them into the energy function. This prefers labelling of object pairs that frequently co-occur in a certain spatial layout and at the same time avoids implausible  labellings during the inference.
Extensive experiments on binary and multi-class segmentation benchmarks demonstrate the promise of the proposed method.
We thus provide new baselines for the segmentation performance on
the Weizmann horse, Graz-02, MSRC-21, Stanford Background and PASCAL VOC 2011 datasets.
\end{abstract}


\end{frontmatter}

\section{Introduction}
\label{sec:intro}
The task of image segmentation is to produce a pixel level labelling of different
object categories, with wide variety of applications ranging from image retrieval to object recognition.
It is challenging as the objects may appear in various backgrounds and
different visual conditions.
\crfs \cite{Lafferty01Conditional}
model the conditional distribution of labels given observations, representing the state-of-the-art
in image/object segmentation  \cite{SzummerKH08,Shotton08,Fulkerson09,Lucchi12,NowozinGL10}.
In \cite{SzummerKH08}, Szummer \etal \ \ proposed to learn the coefficients of \crf potentials using structured support vector machines (\ssvm) and graph cuts. Since then, \ssvm has been widely applied for \crf learning in segmentation tasks.

In the pipeline of \crf learning based image segmentation, finding a good feature representation is of great significance, and can have a profound impact on the segmentation accuracy. Most previous studies rely on hand-crafted features, \eg, using color histograms, HOG or SIFT descriptors to construct bag-of-words features \cite{Fulkerson08,Fulkerson09,Lucchi12,Yao12,Lucchi13}.
Recently, feature learning and especially deep learning methods have gained great popularity
in machine learning and related fields.
This type of methods typically take raw images as input and learn a (deep) representation
of the images, and have found phenomenal success in various tasks such as speech recognition \cite{Hinton06}, image classification
\cite{deepCNN12,decaf13}, object detection \cite{Ross14} \etc See Bengio \etal \ \cite{Bengio13} for a detailed
review.
Deep learning methods attempt to model high-level abstractions in data at multiple layers,
inspired from the cognitive processes of human brains, which generally starts from simpler
concepts to more abstract ones.
The learning is achieved by using deep architectures, \eg, deep belief networks (DBNs) \cite{Hinton06}, stacked autoassociator networks \cite{Bengio07},
deep convolutional
neural networks (CNNs)
\cite{Lecun98,deepCNN12,CVPR15b}, \etc
Among them, CNNs are high-capacity machine learning models with a very large number of (typically a
few million)
parameters that are optimized from labelled training examples.
The success of CNNs in various vision tasks \cite{Lecun98,deepCNN12} is mainly
due to their ability to learn rich mid-level features that accommodate within-class
variance and at the same time possess discriminative information.
This is in  contrast to low-level hand-crafted features.

On the other hand, prior work \cite{Rabinovich07,Ladicky13,Roy14} has demonstrated that holistic reasoning about the occurrences of all classes helps to improve segmentation performance.
These are based on the considerations that neighbouring image regions may be occupied by frequently co-occurring objects, and object pairs of mutual exclusion are less likely to appear together.
For example, a cow is more likely to show up together with grass rather than a monitor, and grass is less likely to appear above sky.
Therefore, we here propose to construct spatially related co-occurrence pairwise potentials to exploit the context information during inference.

In summary, we highlight the main contributions of this work as follows.
\begin{itemize}
\item
We show that cross-domain image features learned by CNNs with labelled data from
ImageNet\footnote{\url{http://image-net.org}} can be successfully transferred for segmentation purpose.
By thoroughly evaluating the performance of the CNN features of different depths and comparing with the traditional bag-of-words and unsupervised feature learning methods, we demonstrate the power of CNN features in image segmentation.
\item
We illustrate that \ssvm based \crf learning with CNN features yields astounding results and thus provide new baselines for segmentation performance on the Weizmann horse, Graz02, MSRC-21, Stanford Background and PASCAL VOC 2011 datasets.
\item
We incorporate spatially related co-occurrence pairwise potentials into the inference and gain further performance boost.

\end{itemize}

\section{Related work}
\begin{figure} [!t]
\centering
    \includegraphics[width=.96\linewidth]{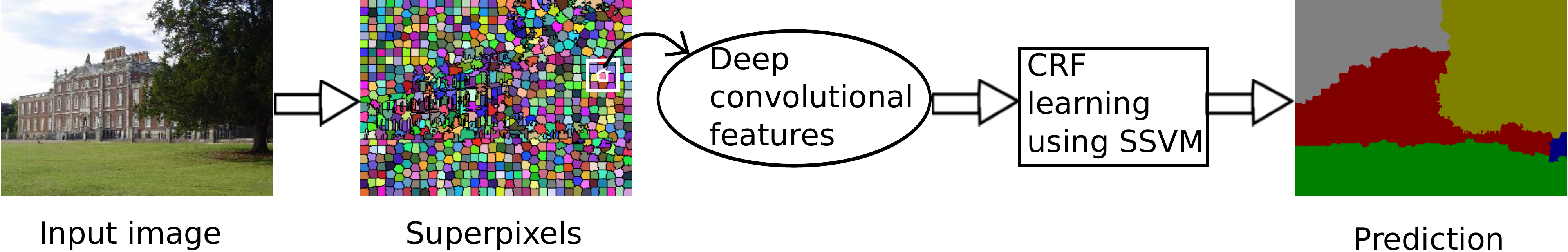}\\
\caption{An illustration of the proposed segmentation pipeline.
We first over-segment the image into superpixels and then compute deep convolutional features of the patch around each superpixel centroid using a pre-trained deep CNN. The learned features are then used to learn a \crf for segmentation.  }
\label{fig:illu_seg}
\end{figure}

We briefly review some work that is relevant to ours.
The first work on using convolutional networks for scene parsing is \cite{Grangier09}.
In \cite{Grangier09}, they train a deep CNN using a supervised greedy learning strategy taking pixels as input to yield a pixel-wise labelling of an image. While somewhat preliminary, they achieved marginal improvement over \crf learning based segmentation methods. We show in this paper that deep CNN features transferred from ImageNet (ImageNet is an image dataset organized according to the WordNet hierarchy, containing millions of labelled images.) combined with \ssvm based \crf learning outperforms most state-of-the-art methods.
Schulz \etal \ \cite{Schulz12} propose to predict the segmentation mask by adding a pairwise class location filter to the conventional CNN architecture of \cite{Lecun98}. In the work of \cite{Lecun13}, the authors use a multiscale convolutional network trained from raw pixels to extract dense feature vectors that encode regions of multiple sizes centered on each pixel and present impressive results on several datasets.
Our work differs from \cite{Lecun13} in two aspects. First, we transfer a deep CNN trained on the ImageNet \cite{deepCNN12} dataset  to segmentation while \cite{Lecun13} trains a 3-stage convolutional network \cite{Lecun98} on the current training data of the segmentation dataset, and we demonstrate experimentally that better performance can be achieved by our method.
Secondly, our method uses \ssvm to learn \crf potentials while no learning is involved in \cite{Lecun13}.
Figure~\ref{fig:illu_seg} shows a sketch of our segmentation pipeline.

Most recently, Girshick \etal \ \cite{Ross14} demonstrate that a deep CNN trained on ImageNet can be successfully transferred to object detection and great performance boost is achieved on the PASCAL VOC 2012 dataset. As an extension of their statement, they also conduct a scene labelling experiment on the PASCAL VOC segmentation dataset to validate the power of deep CNN features on the segmentation task.
Our work is mainly inspired from theirs, but differs in that we combine deep CNN features with \ssvm based \crf learning in contrast to their region proposals and support vector regression based method.  Furthermore, we thoroughly evaluate the performance of deep CNN features compared to the bag-of-words features and unsupervised learned features, and provides new baselines for labelling performance on various segmentation benchmarks.

Co-occurrence statistics have been exploited and demonstrated its strength in the community. In \cite{Rabinovich07}, the authors incorporate semantic object context as a post-processing step by considering the co-occurrence counts of label pairs.
Ladicky \etal \ \cite{Ladicky13} explores the inference methods for \crf with co-occurrence statistics by considering a class of global potentials.
Different from their methods that ignore spatial relations of the co-occurrences, we propose to construct spatially related co-occurrence pairwise potentials, which favor labellings of object pairs that frequently co-occur in a certain spatial layout while at the same time prevents unreasonable labellings.
Our method is  inspired from \cite{Roy14} but differs in that they incorporate the mutex information by adding a mutex constraint to the inference problem while we simply construct co-occurrence pairwise potentials, and most importantly, we explore CNN features combined with \ssvm based \crf learning.

\section{Method}

\subsection{Segmentation using \crf models}

Given $\X=\{\x_i\}$ a collection of image instances with corresponding labels $\Y=\{\y_i\}$, where $i$ indexes images,
\crf \cite{Lafferty01Conditional} considers the log-loss of the overall energy
\begin{align}  \label{eq:crf_log}
P(\y|\x; \w) = \frac{1}{Z} \exp (-\sum_i E(\y_i, \x_i;\w)).
\end{align}
where  $\w$ are parameters and $Z$ the normalization term.
The energy $E$ of an image $\x$ with segmentation labels $\y$ over the nodes (superpixels) $\cal N$ and edges $\cal S$, takes the following form:
\begin{align}
\label{eq:seg_energy}
	E(\y, \x; \w)& =\sum_{p \in {\cal N} } \bh^{(1)}(y^{p}, \x;\w)
	 + \sum_{(p,q) \in {\cal S}} \bh^{(2)}(y^{p}, y^{q}, \x; \w).
\end{align}
Here $ \bx \in {\cal X}, \by \in {\cal Y}$;
$\bh^{(1)}$ and $\bh^{(2)}$ are the unary and pairwise potentials, both of which depend on the observations as well as the parameter $ \w $.
\crf seeks an optimal labelling that achieves maximum a posterior (MAP), which mainly involves a two-step process \cite{SzummerKH08}: 1) Learning the model parameters from the training data; 2) Inferring a most likely label for the test data given the learned parameters.
The segmentation problem thus reduced to minimizing the energy (or cost) over $\y$ by the learned parameters $ \w $, which is:
\begin{align}
\y^{*}=\argmin_{ \y \in  {\cal Y}} E(\y, \x; \w).
\end{align}

\subsection{Learning \crf in the large-margin framework}
Applying the large-margin based \crf learning is to solve the following optimization:
\begin{align}
  \min_{\bw, \bxi \geq 0}   \;\; &
        \half \fnorm{\bw}^2 +  {\tfrac{C}{m}} \, \sum_i \xi_i \notag \\
  \st  \;\; &
    E(\y, \x_i; \w)- E (\y_i, \x_i; \w )\geq
    \loss ( \y_i, \y) - \xi_i, \notag \\
   & \quad\quad\quad
   \forall i=1, \dots, m,
   \text{ and }\forall \y \in  {\cal Y};.
\label{eq:ssvm_crf}
\end{align}
where $\loss: \calY \times \calY \mapsto \Real$ is a loss function associated with the prediction and the true label mask.  In general, we have $ \loss( \y, \y ) = 0$ and $ \loss(\y, \z ) > 0 $ for any $ \z \neq \y $.
Intuitively, the optimization in \eqref{eq:ssvm_crf} is to encourage the
energy of the ground truth label $E(\y_i,\x_i; \w)$ to be lower than any other
{\em incorrect} labels $E(\y, \x_i; \w)$ by at least a margin $\loss (
\y_i, \y)$. The \ssvm solves \eqref{eq:ssvm_crf} by iteratively finding the most violated constraint for each example $i$:
\begin{align}
\y_i^{*}=\argmin_{ \y \in  {\cal Y}} E(\y, \x; \w) - \loss(\y_i, \y).
\label{eq:inf_crf}
\end{align}

To learn \crf in the large margin framework, we consider energy functions that are linear in the parameter $\w$, which indicates that the unary and the pairwise potentials in \eqref{eq:seg_energy} can be written as:
\begin{align}
\bh^{(1)}(y^{p}, \x;\w) = \left\langle\w^{(1)}, \phi^{(1)}(y^p, \x)\right\rangle,
\end{align}
and
\begin{align}
\bh^{(2)}(y^{p}, y^{q}, \x;\w) = \left\langle \w^{(2)}, \phi^{(2)}(y^{p}, y^{q}, \x) \right\rangle,
\end{align}
where $\phi^{(1)}, \phi^{(2)}$ are the unary and pairwise feature mappings respectively and $\left\langle \cdot, \cdot \right\rangle$ denotes inner products. Clearly we have $\w=\w^{(1)}\otimes\w^{(2)}$ ($\otimes$ stacks two vectors).
We will show how to construct the feature mappings over the learned deep features in the following.
\paragraph{\bf{Implementation details}}
After obtaining the learned deep features, we define feature mappings upon them to construct the energy function. Consider the image $\x$ with label $\y$, let $\x^p$ be the feature vector associated with the $p$-th superpixel, and $K$ is the number of classes (possible labels). Then we define the unary feature mappings as
\begin{align}
\phi^{(1)}(y^p, \x)=[I(y^p=1)\x^{p \T}, \ldots, I(y^p=K)\x^{p\T}]^\T,
\end{align}
where $I(\cdot)$ is an indicator function which equals $1$ if the input is true and $0$ otherwise.
In the case of multi-class, the dimension of $\phi^{(1)}(y^p, \x)$ can be too large when $\x^p$ is high dimensional. To address this issue, we first train an one-vs-all multi-class linear SVM over the features of superpixels, and then use the output confidence scores of the p-th superpixel as $\x^p$ to construct the unary potential.
Similar strategy is used in \cite{Fulkerson09,Lucchi12}.
Accordingly, the pairwise feature mapping is constructed as
\begin{align}
\label{eq:pairwise_feat}
\phi^{(2)}(y^p, y^q, \x)= L_{pq} \cdot I(y^p \neq y^q),
\end{align}
where $L_{pq}$ can be the shared boundary length or  inversed color difference between neighbouring superpixels.

The energy function in \eqref{eq:seg_energy} can then be written as
\begin{align}
\label{eq:energy}
	E(\y, \x; \w)& = \left\langle \w^{(1)},  \sum_{p \in {\cal N} } \phi^{(1)}(y^{p}, \x) \right \rangle
	 +  \left\langle \w^{(2)}, \sum_{(p,q) \in {\cal S}} \phi^{(2)}(y^{p}, y^{q}, \x) \right\rangle.
\end{align}

To deal with the unbalanced appearance of different categories in the dataset, we define $\loss ( \y_i, \y)$ as the weighted Hamming loss, which weighs errors for a given class inversely proportional to the frequency it appears in the training data, similar to \cite{Lucchi12}.
We use the method of \cite{Zhangzhen13} to solve the inference in \eqref{eq:inf_crf}.

\subsection{Inference with co-occurrence pairwise potentials}
\label{sec:co-occur}
To fully exploit context information, we consider the frequency of co-occurred object pairs in different spatial layouts during the inference.
On one hand, this prefers labelling of frequently co-occurred label pairs  in a certain spatial relation; while on the other hand, it excludes unreasonable labellings of co-occurrences (mutex constraint, similar as \cite{Roy14}), such as grass, water or road appearing above sky.
Different from the mutex constraint used in \cite{Roy14}, we incorporate the co-occurrence constraint into the pairwise term by devising spatially related co-occurrence pairwise potentials.
We consider four spatial relations of the adjacent superpixel pairs: p is above q, p is below q, p is left to q and p is right to q. Then the feature mapping for the pairwise potential in \eqref{eq:energy} is written as:
\begin{align}
\label{eq:pairwise}
 \sum_{(p,q) \in {\cal S}} \phi^{(2)}(y^{p}, y^{q}, \x)   =  \sum_{(p,q) \in {\cal S}_1} \phi_1^{(2)} (y^p, y^q, \x)  + \sum_{(p,q) \in {\cal S}_2} \phi_2^{(2)} (y^p, y^q, \x) \nonumber \\
+ \sum_{(p,q) \in {\cal S}_3} \phi_3^{(2)} (y^p, y^q, \x)  + \sum_{(p,q) \in {\cal S}_4} \phi_4^{(2)} (y^p, y^q, \x) .
\end{align}
where
${\cal S}_1$, ${\cal S}_2$, ${\cal S}_3$, ${\cal S}_4$ are the sets of edges where p and q are in the spatial relations ``above'', ``below'', ``left'' and ``right'' respectively, and ${\cal S} = {\cal S}_1 \cup {\cal S}_2 \cup {\cal S}_3 \cup {\cal S}_4$, and ${\cal S}_i \cap {\cal S}_j=\emptyset$ for $i \neq j, i,j=1,2,3,4$.

To construct the co-occurrence pairwise potentials, we assume that the training data is sufficiently large.
The pairwise potentials in \eqref{eq:pairwise} can then be written as:
\begin{align}
\phi_i^{(2)}(y^p, y^q, \x)= L_{pq} \cdot I(y^p \neq y^q) \cdot g_i(y^p, y^q), i=1,2,3,4.
\end{align}
where $g_i(y^p, y^q) = \frac{1}{f^i_{co-occur}(y^p, y^q)}$ with $f^i_{co-occur}(y^p, y^q)=\frac{N^i_{pq}}{N_{pq}}$. Here, $N_{pq}$ is the number of training images in which $y^p$ and $y^q$ co-exist, and $N^i_{pq}$ ($i=1,2,3,4$) are the numbers of training images in which $y^p$ and $y^q$ appear in the four spatially related neighbouring superpixels respectively.
If $N^i_{pq}=0$, meaning that  $y^p$ and $y^q$  never appear in the $i$th spatial relation, then $g_i(y^p, y^q)=inf$, preventing the inference to yield such pair labellings.
Intuitively, this would prefer labellings that frequently co-occurred in certain spatial relations in the training data, and  avoid those mutual exclusion labellings,  such as grass appear above sky.

Note that the mutex constraint used in \cite{Roy14} can be seen a special case of our co-occurrence pairwise potentials, as it is equivalent to ours when we set $g_i(y^p, y^q)=inf$  for  $f^i_{co-occur}(y^p, y^q)=0$ and $g_i(y^p, y^q)=1$  for  $f^i_{co-occur}(y^p, y^q) \neq 0$.
We will provide experimental comparison with this case in Section~\ref{sec:exp2}.
After learning the \crf using \ssvm, we construct co-occurrence pairwise potentials for prediction.
We add a trade-off parameter $\alpha$ multiplied to the pairwise term and tune it from 0.5 to 2 based on validation sets.

\section{Experiments}
To demonstrate the effectiveness of the proposed method, we first compare the CNN features with the traditional bag-of-words feature and an unsupervised feature learning method \cite{Coates11} as well as evaluate the impact of depths to the performance of the CNN features in Section~\ref{sec:exp1}.
We then compare with state-of-the-art methods on several image segmentation datasets in Section~\ref{sec:exp2}.

\subsection{Experimental setup}
For the CNN features, we use the model trained on ImageNet provided by Caffe \cite{Jia13caffe}.
The network follows the famous AlexNet \cite{deepCNN12}, and is composed of $5$ convolutional layers and $2$ fully connected layers together with a soft-max layer.

We evaluate the performance of the proposed method on Weizmann horse, Graz-02,  MSRC-21, Standford Background and PASCAL VOC 2011 segmentation challenge dataset.
The Weizmann horse dataset\footnote{\url{ http://www.msri.org/people/members/eranb/} }  consists of 328 horse images from various backgrounds, with groundtruth masks available for each image. We use the same data split as in \cite{Bertelli11}, \cite{Kuettel12}, and we simply resize the images to $256 \times 256$.
The Graz-02
dataset\footnote{\url{http://www.emt.tugraz.at/~pinz/} } contains 3 categories (bike, car and people). 
This dataset is considered challenging as the objects appear at various background and with different poses.  
We follow the evaluation protocol in \cite{Marszalek07} to use 150 for training and 150 for testing for each category.

The MSRC-21 dataset \cite{Shotton08} is a popular multi-class segmentation benchmark with 591 images containing objects from 21 categories. We follow the standard split to divide the dataset into training/validation/test subsets. 
The Standford Background dataset \cite{Gould09} is a collection of outdoor scene images from several publicly available datasets,  which consists of 715 images coming from 8 categories. Each image is approximately $320 \times 240$ pixels and contains at least one foreground object. We use the same evaluation protocol as in \cite{Gould09} to report 5-fold cross validation accuracy (global and per-category).
The VOC 2011 dataset consists of images from 20 objects and background.
We train on the training set and test on the validation images. 
The performance are quantified by the standard VOC measure \cite{PascalVOC}.

We start with over-segmenting the images into superpixels using SLIC \cite{slic} ($\sim 700$ superpixels per image) and then compute features  within regions around each superpixel centroid with different block sizes ($36 \times 36$, $48 \times 48$, $64 \times 64$, $72 \times 72$ ). 
We construct four types of pairwise features also using different block sizes to enforce spatial smoothness, which are color difference in LUV space, color histogram difference, texture difference in terms of LBP operators as well as shared boundary length \cite{Fulkerson09}. 
Training our model on the MSRC-21 dataset takes around 2 hours.
During prediction, the inference is rather efficient (less than 1 sec per image).

\subsection{Baseline Comparison}
\label{sec:exp1}
To show the superiority of the deep CNN over the unsupervised feature learning, we compare with the traditional bag-of-word (BoW) feature and features learned from a popular unsupervised feature learning method \cite{Coates11}. Specifically, we first extract dense SIFT descriptors within each superpixel block and then quantize them into BoW feature using nearest neighbour search with a codebook size of 400. 
For the unsupervised feature learning, we first learn a dictionary of size 400 and patch size 6$\times$6 based on the evaluated image dataset using Kmeans, and then use the soft threshold coding \cite{Coates11} %
to encode patches extracted from each superpixel block. The final feature vectors are obtained by performing a three-level max pooling over the superpixel block. 

To investigate the roles of different layers in the proposed segmentation method, we evaluate the performance of features from the last three layers of the CNN model (5th, 6th and 7th layers). The 5th layer (with dimension 9216) is the last convolutional layer of the CNN. 
The 6th layer (with dimension 4096) is a fully connected layer follows the 5th layer and the 7th (with dimension 4096) is the final layer of the feature learning pipeline.
Using the two types of learned features, we compare the \ssvm based \crf learning with a baseline method, namely, linear \svm, which classifies each superpixel independently without \crf learning.
The datasets used in this section are Weizmann horse, Graz-02 and MSRC-21.
We use BoW to denote the bag-of-words feature, UFL represent the unsupervised feature learning method, and L5, L6, L7 are CNN features of the 5th, 6th and 7th layer respectively.

\begin{figure} [t]
\centering
    \includegraphics[width=.11\linewidth, height=0.45in]{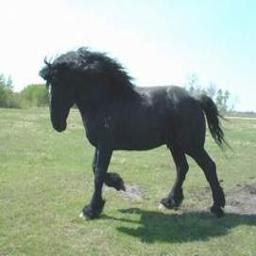}
     \includegraphics[width=.11\linewidth, height=0.45in]{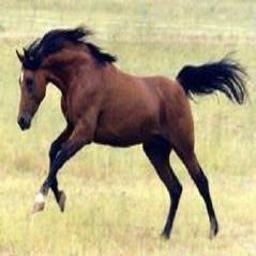}
     \includegraphics[width=.11\linewidth, height=0.45in]{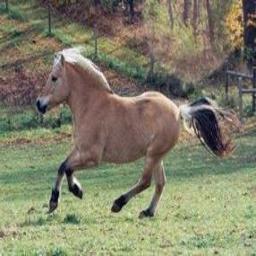}       
	\includegraphics[width=.11\linewidth, height=0.45in]{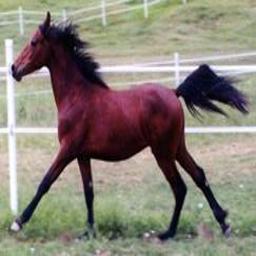}                  
    \includegraphics[width=.11\linewidth, height=0.45in]{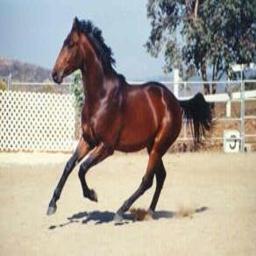}
     \includegraphics[width=.11\linewidth, height=0.45in]{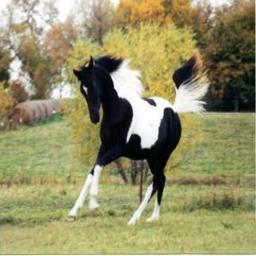}  
     \includegraphics[width=.11\linewidth, height=0.45in]{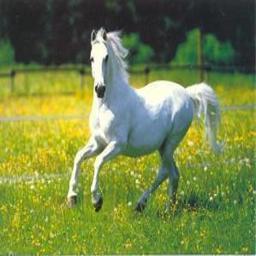}   
     \includegraphics[width=.11\linewidth, height=0.45in]{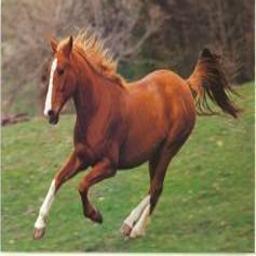} \\
     
     \includegraphics[width=.11\linewidth, height=0.45in]{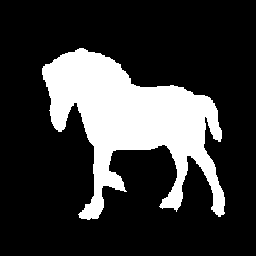}
     \includegraphics[width=.11\linewidth, height=0.45in]{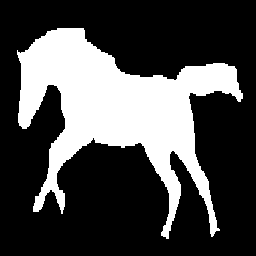}
     \includegraphics[width=.11\linewidth, height=0.45in]{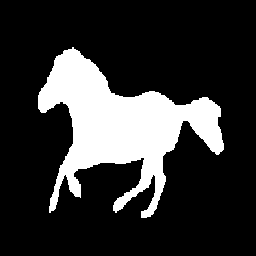}       
	\includegraphics[width=.11\linewidth, height=0.45in]{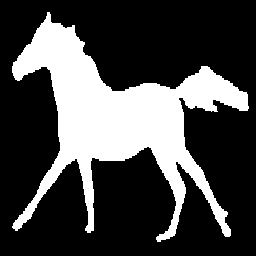}                  
    \includegraphics[width=.11\linewidth, height=0.45in]{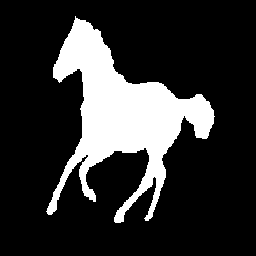}
     \includegraphics[width=.11\linewidth, height=0.45in]{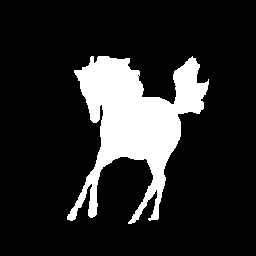}  
     \includegraphics[width=.11\linewidth, height=0.45in]{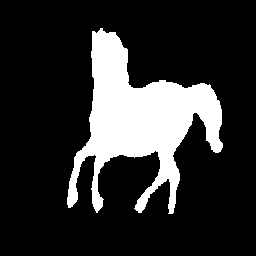}   
     \includegraphics[width=.11\linewidth, height=0.45in]{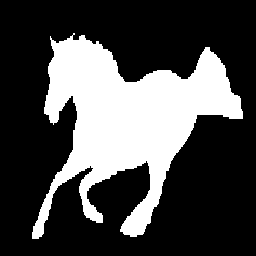} \\

     \includegraphics[width=.11\linewidth, height=0.45in]{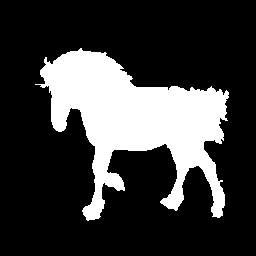}
     \includegraphics[width=.11\linewidth, height=0.45in]{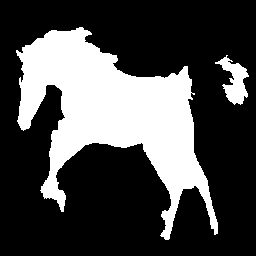}
     \includegraphics[width=.11\linewidth, height=0.45in]{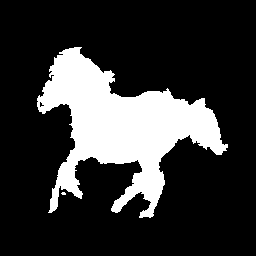}       
	\includegraphics[width=.11\linewidth, height=0.45in]{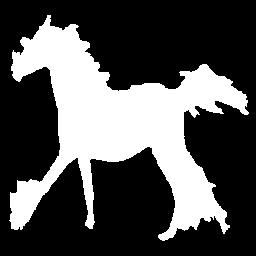}                  
    \includegraphics[width=.11\linewidth, height=0.45in]{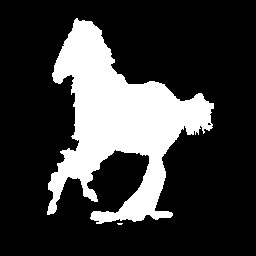}
     \includegraphics[width=.11\linewidth, height=0.45in]{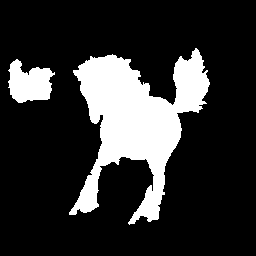}  
     \includegraphics[width=.11\linewidth, height=0.45in]{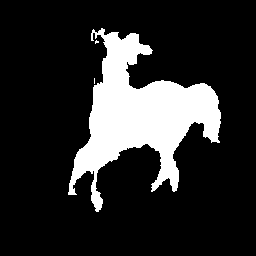}   
     \includegraphics[width=.11\linewidth, height=0.45in]{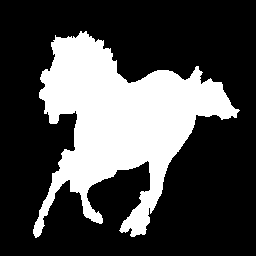} \\
     
     \includegraphics[width=.11\linewidth, height=0.45in]{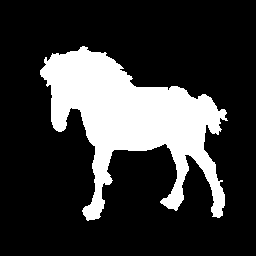}
     \includegraphics[width=.11\linewidth, height=0.45in]{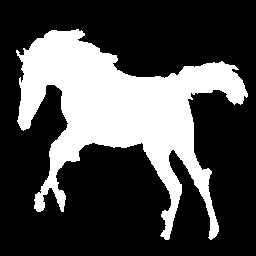}
     \includegraphics[width=.11\linewidth, height=0.45in]{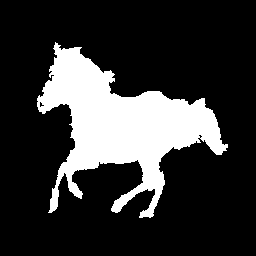}       
	\includegraphics[width=.11\linewidth, height=0.45in]{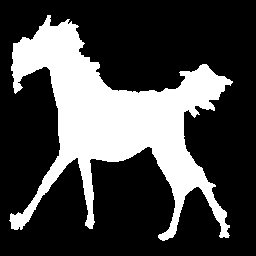}                  
    \includegraphics[width=.11\linewidth, height=0.45in]{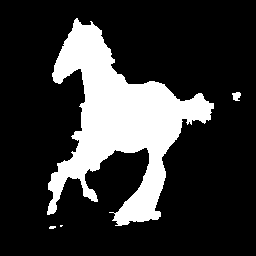}
     \includegraphics[width=.11\linewidth, height=0.45in]{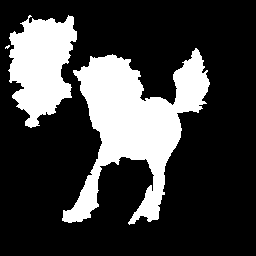}  
     \includegraphics[width=.11\linewidth, height=0.45in]{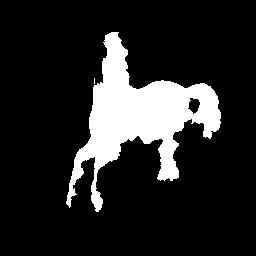}   
     \includegraphics[width=.11\linewidth, height=0.45in]{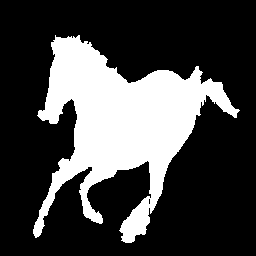} \\
     
     \includegraphics[width=.11\linewidth, height=0.45in]{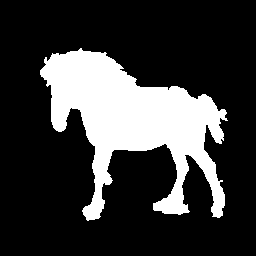}
     \includegraphics[width=.11\linewidth, height=0.45in]{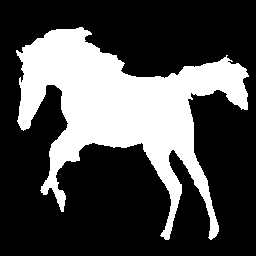}
     \includegraphics[width=.11\linewidth, height=0.45in]{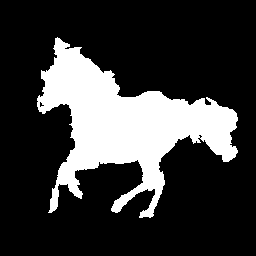}       
	\includegraphics[width=.11\linewidth, height=0.45in]{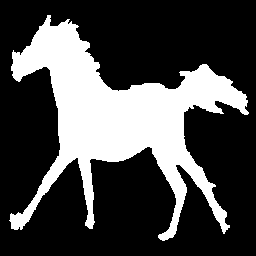}                  
    \includegraphics[width=.11\linewidth, height=0.45in]{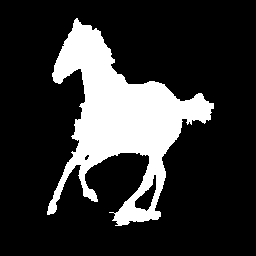}
     \includegraphics[width=.11\linewidth, height=0.45in]{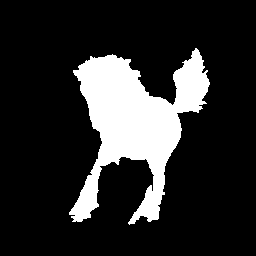}  
     \includegraphics[width=.11\linewidth, height=0.45in]{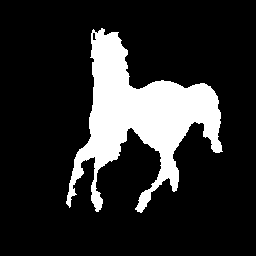}   
     \includegraphics[width=.11\linewidth, height=0.45in]{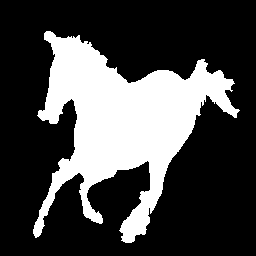} \\

\caption{Segmentation examples on Weizmann horse. 1st row: Test images; 2nd row: Ground truth; 
3rd row: Segmentation results produced by \ssvm based \crf learning with bag-of-words feature;
4th row: Segmentation results produced by \ssvm based \crf learning with unsupervised feature learning;
5th row: Segmentation results produced by \ssvm based \crf learning with the 6th layer CNN features. }
\label{fig:seg_horse}     
\end{figure}

\paragraph{Weizmann horse}
We first test on the Weizmann horse dataset.
The performance are quantified by the global pixel-wise accuracy $S_a$ and the foreground intersection over union score $S_o$, similar as in \cite{Bertelli11}. $S_a$ measures the percentage of pixels correctly classified while $S_o$ directly reflects the segmentation quality of the foreground. The compared results are reported in Table~\ref{tab:seg_horse_baseline}.
We can observe that the CNN features perform consistently better than the bag-of-words feature and the unsupervised learned feature in both \svm and \ssvm.
By enforcing smoothness term, \ssvm based \crf learning obtain far better segmentations than simple binary model as \svm.
Furthermore, features of different depths exhibit almost similar performance with the 6th layer performs marginally better than the other compared layers in both \svm and \ssvm.
In Figure~\ref{fig:seg_horse}, we show some examples of qualitative evaluation, which yields conclusions that are in accordance with those from Table~\ref{tab:seg_horse_baseline}.

\begin{table}[t]
\centering
\resizebox{0.65\linewidth}{!} {
\begin{tabular}{| c |c c c c c | c c  c c c| }
\hline
\multirow{2}{*}{Metric} &\multicolumn{5}{c|}{\svm}  &\multicolumn{5}{c|}{\ssvm} \\
\cline{2-11}
 &BoW &UFL &L5 &L6 &L7  &BoW &UFL &L5 &L6 &L7 \\
\hline
Sa &87.5 &89.3 &90.1 &\textbf{92.7}	&91.1  &92.3 &94.6 &95.2	 &\textbf{95.7}	&95.1 \\
So &58.7 &63.6 &68.9  &\textbf{74.6} &72.9 &72.5 &80.1 &82.4  &\textbf{84.0}  &82.3	\\ 
\hline
\end{tabular}
}
\caption{Performance of different methods on the Weizmann horse dataset.
CNN features perform significantly better than the traditional BoW feature and the unsupervised feature learning method, with features of the 6th layer performing marginally better than other compared layers. \ssvm based \crf learning performs far better than \svm.}
\label{tab:seg_horse_baseline}
\end{table}  

\begin{figure} [t]
\centering
    \includegraphics[width=.1\linewidth, height=0.45in]{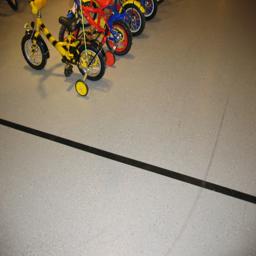}
     \includegraphics[width=.1\linewidth, height=0.45in]{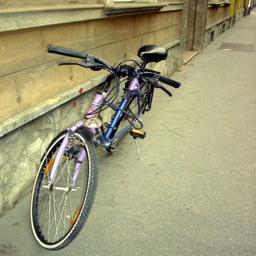}
     \includegraphics[width=.1\linewidth, height=0.45in]{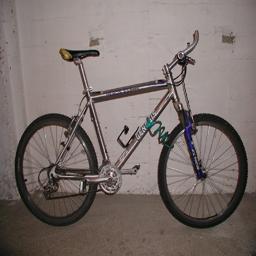}       
	\includegraphics[width=.1\linewidth, height=0.45in]{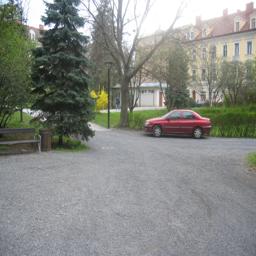}                  
    \includegraphics[width=.1\linewidth, height=0.45in]{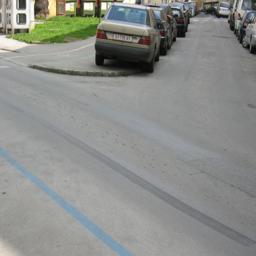}
     \includegraphics[width=.1\linewidth, height=0.45in]{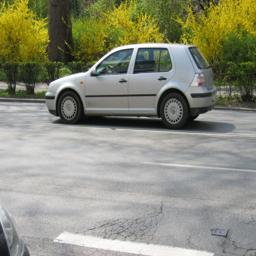}  
     \includegraphics[width=.1\linewidth, height=0.45in]{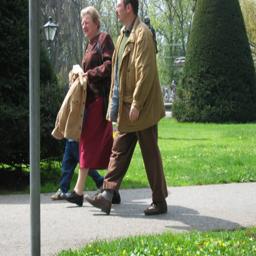}   
     \includegraphics[width=.1\linewidth, height=0.45in]{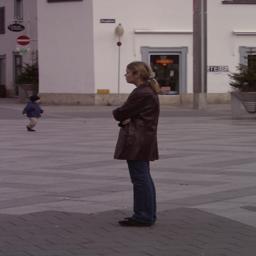} 
     \includegraphics[width=.1\linewidth, height=0.45in]{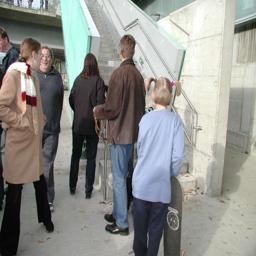} \\
     
     \includegraphics[width=.1\linewidth, height=0.45in]{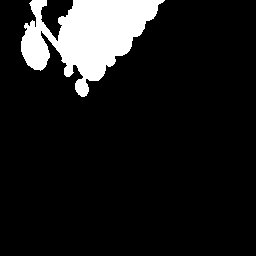}
     \includegraphics[width=.1\linewidth, height=0.45in]{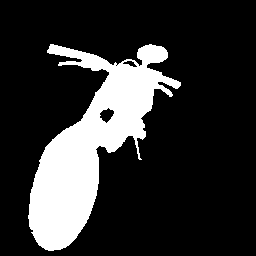}
     \includegraphics[width=.1\linewidth, height=0.45in]{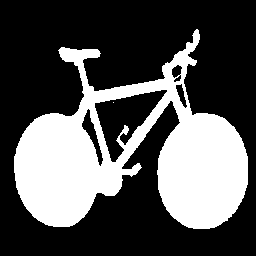}       
	\includegraphics[width=.1\linewidth, height=0.45in]{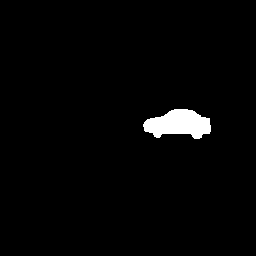}                  
    \includegraphics[width=.1\linewidth, height=0.45in]{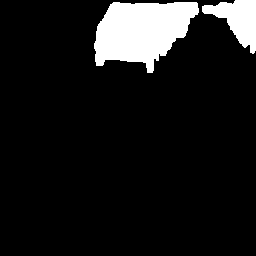}
     \includegraphics[width=.1\linewidth, height=0.45in]{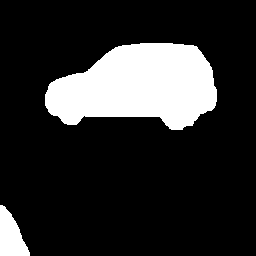}  
     \includegraphics[width=.1\linewidth, height=0.45in]{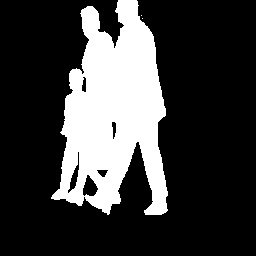}   
     \includegraphics[width=.1\linewidth, height=0.45in]{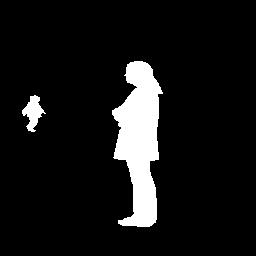} 
     \includegraphics[width=.1\linewidth, height=0.45in]{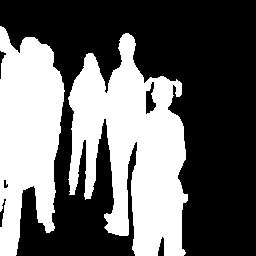} \\

     \includegraphics[width=.1\linewidth, height=0.45in]{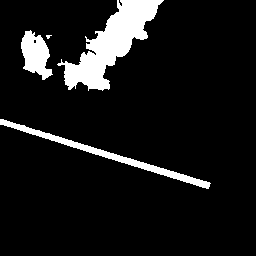}
     \includegraphics[width=.1\linewidth, height=0.45in]{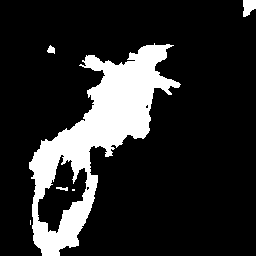}
     \includegraphics[width=.1\linewidth, height=0.45in]{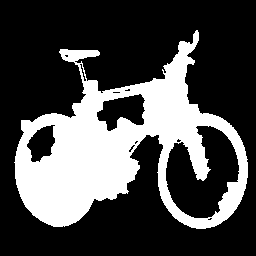}       
	\includegraphics[width=.1\linewidth, height=0.45in]{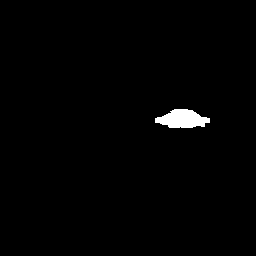}                  
    \includegraphics[width=.1\linewidth, height=0.45in]{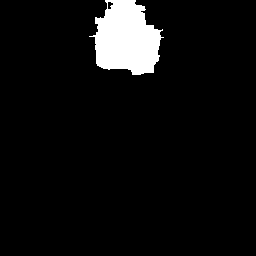}
     \includegraphics[width=.1\linewidth, height=0.45in]{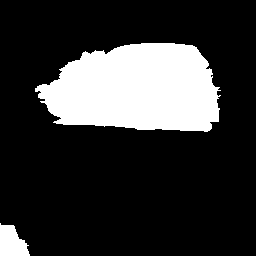}  
     \includegraphics[width=.1\linewidth, height=0.45in]{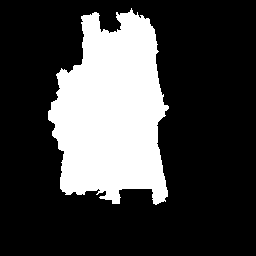}   
     \includegraphics[width=.1\linewidth, height=0.45in]{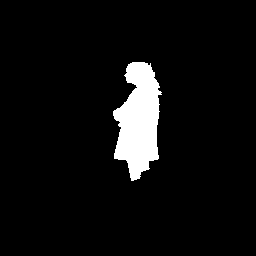} 
     \includegraphics[width=.1\linewidth, height=0.45in]{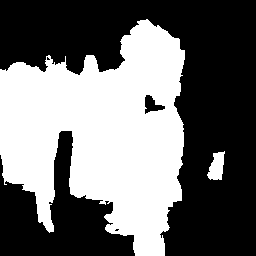} \\
     
     \includegraphics[width=.1\linewidth, height=0.45in]{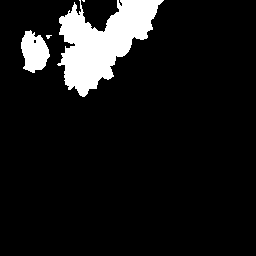}
     \includegraphics[width=.1\linewidth, height=0.45in]{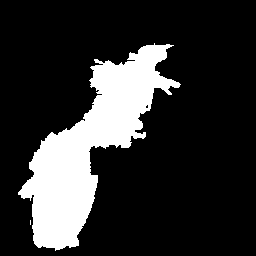}
     \includegraphics[width=.1\linewidth, height=0.45in]{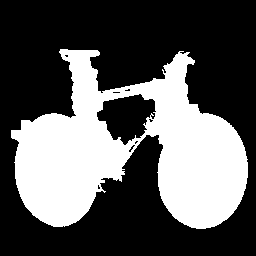}       
	\includegraphics[width=.1\linewidth, height=0.45in]{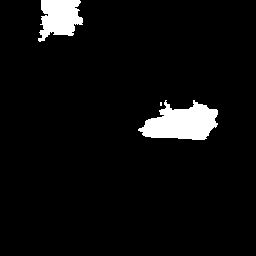}                  
    \includegraphics[width=.1\linewidth, height=0.45in]{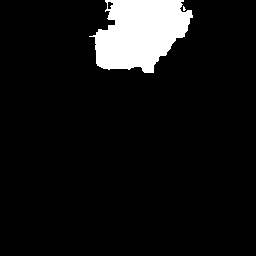}
     \includegraphics[width=.1\linewidth, height=0.45in]{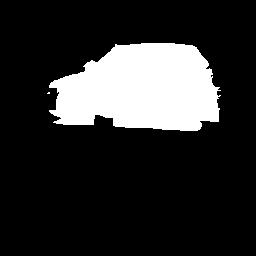}  
     \includegraphics[width=.1\linewidth, height=0.45in]{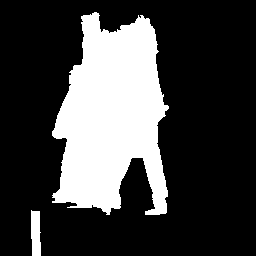}   
     \includegraphics[width=.1\linewidth, height=0.45in]{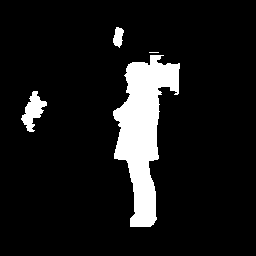} 
     \includegraphics[width=.1\linewidth, height=0.45in]{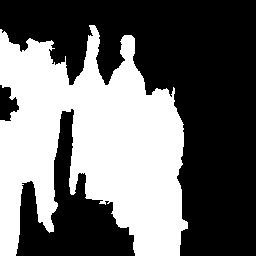} \\

     \includegraphics[width=.1\linewidth, height=0.45in]{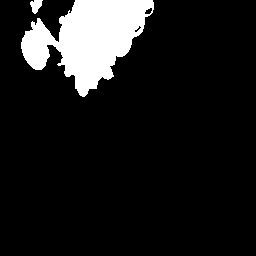}
     \includegraphics[width=.1\linewidth, height=0.45in]{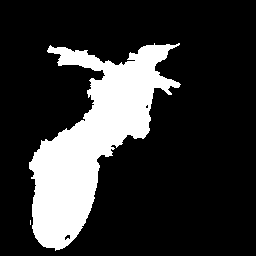}
     \includegraphics[width=.1\linewidth, height=0.45in]{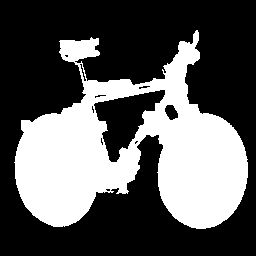}       
	\includegraphics[width=.1\linewidth, height=0.45in]{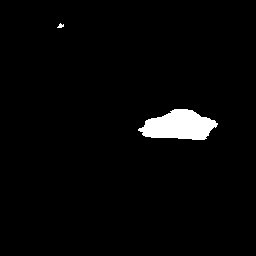}                  
    \includegraphics[width=.1\linewidth, height=0.45in]{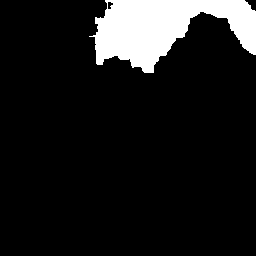}
     \includegraphics[width=.1\linewidth, height=0.45in]{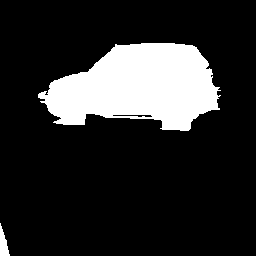}  
     \includegraphics[width=.1\linewidth, height=0.45in]{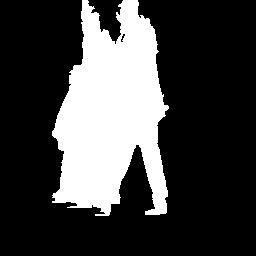}   
     \includegraphics[width=.1\linewidth, height=0.45in]{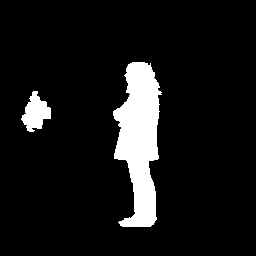} 
     \includegraphics[width=.1\linewidth, height=0.45in]{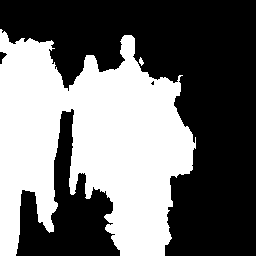} \\

\caption{Segmentation examples on the Graz-02 dataset. 1st row: Test images; 2nd row: Ground truth; 
3rd row: Segmentation results produced by \ssvm based  \crf learning with bag-of-words feature;
4th row: Segmentation results produced by \ssvm based  \crf learning with unsupervised feature learning; 
5th row: Segmentation results produced by \ssvm based  \crf learning with the 6th layer CNN features. }
\label{fig:seg_graz02}     
\end{figure}

\paragraph{Graz-02}
For a comprehensive evaluation, we use two  measurements to quantify the performance of our method on the Graz-02 dataset, which are intersection over union score and the pixel accuracy (including foreground and background). We report the results in Table~\ref{tab:seg_graz02_baseline}.
It can be observed that feature learning methods generally outperform the traditional bag-of-words feature, with CNN features standing as the best.
As for different depths, feature of the 6th layer consistently outperforms all the other compared layers in both \svm and \ssvm, which is in accordance with the conclusion of \cite{Ross14}.
We show some segmentation examples in Figure~\ref{fig:seg_graz02}, from which
we can see that \ssvm based \crf learning with CNN features produces segmentation similar to ground truth. 

\begin{table}[!t]
\centering
\resizebox{1\linewidth}{!}
  {
  \begin{tabular}{|l c | c| c| c |c| c| c |}
    \hline 
    \multicolumn{2}{|c|}{Category} &bike &car &people  &bike &car &people \\  
  \hline 
    & &\multicolumn{3}{c|}{intersection/union (foreground, background) ($\%$)}  
    &\multicolumn{3}{c|}{pixel accuracy (foreground, background) ($\%$)} \\ 
  \hline
  \multirow{5}{*}{\svm} 
   &BoW &66.5 (50.4, 82.7) &66.8 (42.2, 91.5) &64.0 (41.9, 86.2) 
  	   &79.0 (67.9, 90.2) &75.8 (55.2, 96.3) &74.5 (55.4, 93.7)\\
   &UFL &69.7 (55.0, 84.5) &73.1 (52.7, 93.4) &61.4 (37.2, 85.6)
   	  &81.7 (72.4, 91.1) &80.9 (64.4, 97.4) &71.2 (48.2, 94.3)\\  
   &L5 &74.6 (62.4, 86.8) &76.0 (58.4, 93.7) &65.9 (47.0, 84.9)
   	   &86.3 (81.2, 91.4) &86.3 (76.2, 96.4) &80.9 (72.4, 89.4) \\
   &L6 &\textbf{77.7} (66.7, 88.6) &\textbf{78.1} (61.8, 94.5)  &\textbf{68.9} (51.1, 86.6) 
   	    &\textbf{88.4} (84.4, 92.5) &\textbf{87.2} (77.3, 97.0)  &\textbf{83.0} (75.2, 90.8) \\
   &L7  &77.1 (66.0, 88.2)  &77.6 (60.8, 94.3)  &68.4 (50.5, 86.3) 
       &88.2(84.1, 92.2)  &86.6 (76.3, 97.0)  &82.8 (75.1, 90.5) \\
   \hline
   \multirow{5}{*}{\ssvm} 
   &BoW &70.9 (56.6, 85.2) &75.7 (57.2, 94.1) &71.3 (53.5, 89.1)
   	    &82.5 (73.5, 91.6) &83.2 (68.9, 97.6) &81.4 (68.2, 94.7) \\
   &UFL &74.2 (61.5, 86.9) &77.9 (60.9, 94.9) &70.9 (53.0, 88.8)
        &85.4 (78.6, 92.1) &83.8 (69.3, 98.4) &81.5 (68.9, 94.2) \\     
   &L5 &81.6 (72.3, 90.8)   &84.5 (72.6, 96.4) &75.4 (61.1, 89.7) 
   	    &91.0 (88.0, 93.9) &90.6 (82.8, 98.3) &88.8 (85.3, 92.3) \\
   &L6 &\textbf{82.0} (73.1, 91.0) &\textbf{85.6} (74.5, 96.6)  &\textbf{79.6} (67.2, 92.1) 
       &\textbf{91.6} (89.5, 93.7) &\textbf{91.4} (84.4, 98.4)  &\textbf{90.0} (85.1, 94.8)  \\
   &L7  &81.7 (72.6, 90.8)  &85.1 (73.7, 96.5)  &76.0 (62.0, 90.0) 
   	    &91.3 (89.0, 93.6)   &91.2 (84.0, 98.4)  &89.3 (86.1, 92.4) \\
   \hline 
  \end{tabular}
  }   
\caption{Compared results of the average intersection-over-union score and average pixel accuracy on the Graz-02 dataset.
We include the foreground and background results in the brackets. 
CNN features perform significantly better than the traditional BoW feature and the unsupervised feature learning, with features of the 6th layer performing the best among the compared layers in both \svm and \ssvm. \ssvm based \crf learning performs far better than \svm.}
\label{tab:seg_graz02_baseline}
\end{table}

\begin{figure} [!t]
\centering
	\includegraphics[width=.112\linewidth, height=0.45in]{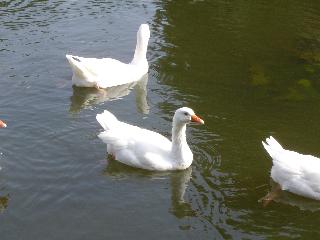}
     \includegraphics[width=.112\linewidth, height=0.45in]{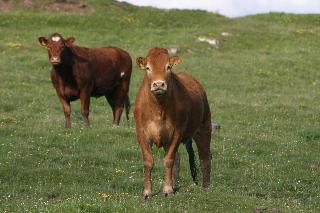}     
	\includegraphics[width=.112\linewidth, height=0.45in]{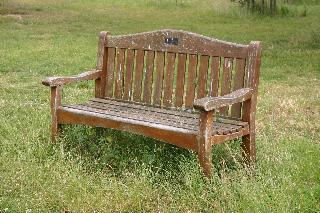}
    \includegraphics[width=.112\linewidth, height=0.45in]{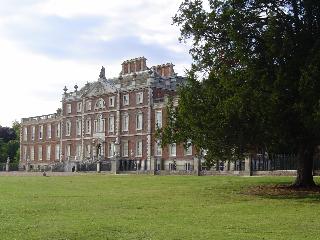}
     \includegraphics[width=.112\linewidth, height=0.45in]{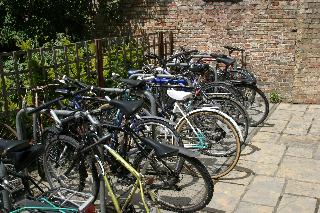}
     \includegraphics[width=.112\linewidth, height=0.45in]{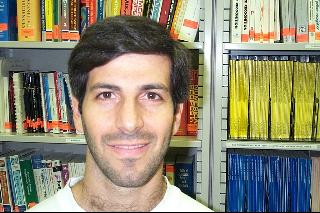}   
     \includegraphics[width=.112\linewidth, height=0.45in]{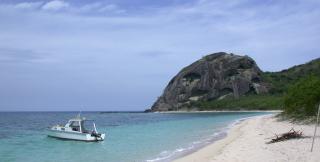} 
     \includegraphics[width=.112\linewidth, height=0.45in]{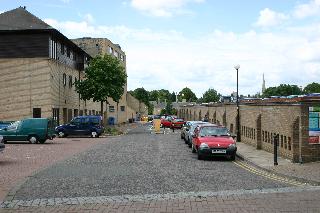}\\

	\includegraphics[width=.112\linewidth, height=0.45in]{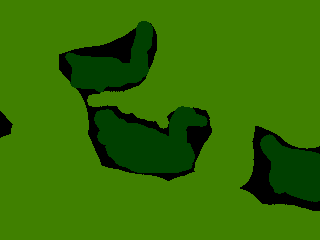}
     \includegraphics[width=.112\linewidth, height=0.45in]{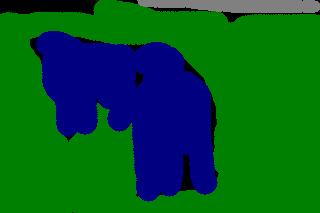}     
	\includegraphics[width=.112\linewidth, height=0.45in]{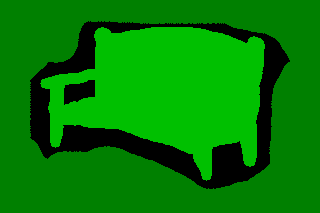}
	\includegraphics[width=.112\linewidth, height=0.45in]{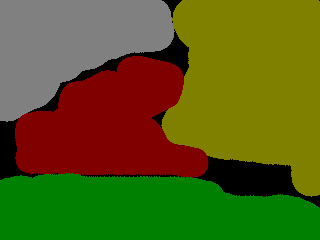}
     \includegraphics[width=.112\linewidth, height=0.45in]{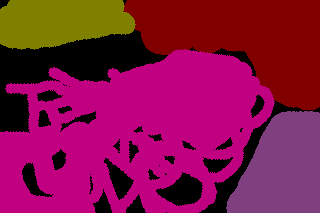}
     \includegraphics[width=.112\linewidth, height=0.45in]{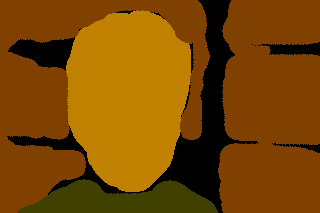}   
     \includegraphics[width=.112\linewidth, height=0.45in]{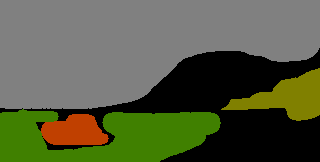} 
     \includegraphics[width=.112\linewidth, height=0.45in]{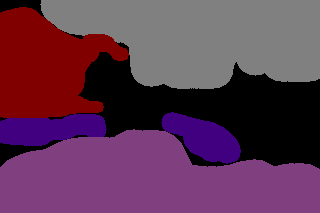}\\

	\includegraphics[width=.112\linewidth, height=0.45in]{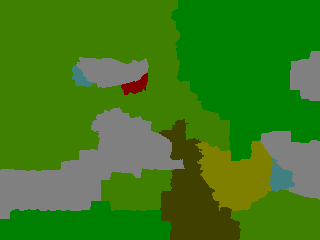}
     \includegraphics[width=.112\linewidth, height=0.45in]{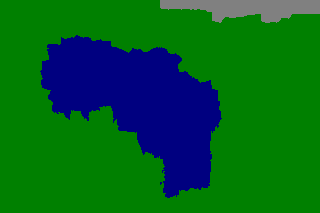}     
	\includegraphics[width=.112\linewidth, height=0.45in]{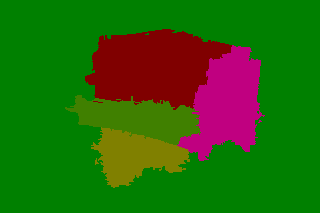}
    \includegraphics[width=.112\linewidth, height=0.45in]{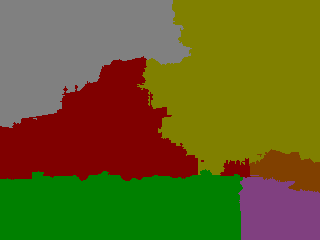}
     \includegraphics[width=.112\linewidth, height=0.45in]{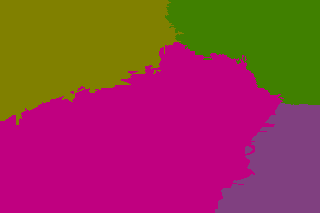}
     \includegraphics[width=.112\linewidth, height=0.45in]{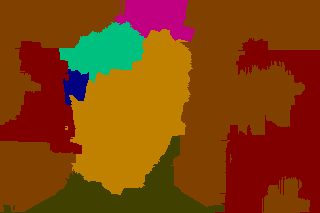}   
     \includegraphics[width=.112\linewidth, height=0.45in]{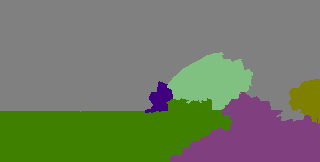} 
     \includegraphics[width=.112\linewidth, height=0.45in]{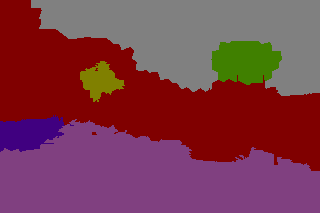}\\

	\includegraphics[width=.112\linewidth, height=0.45in]{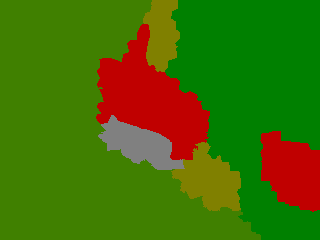}
     \includegraphics[width=.112\linewidth, height=0.45in]{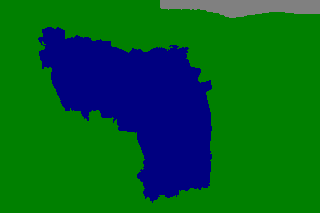}     
	\includegraphics[width=.112\linewidth, height=0.45in]{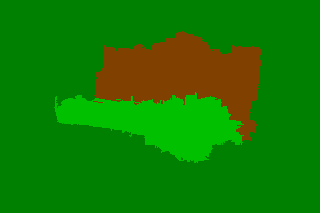}
    \includegraphics[width=.112\linewidth, height=0.45in]{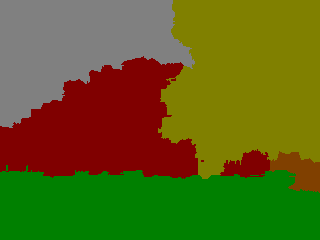}
     \includegraphics[width=.112\linewidth, height=0.45in]{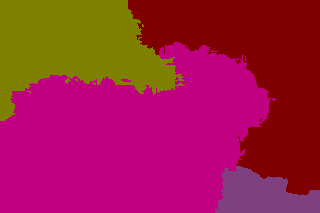}
     \includegraphics[width=.112\linewidth, height=0.45in]{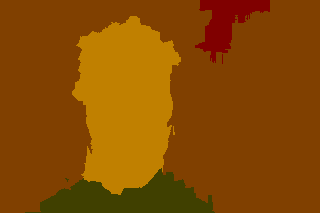}   
     \includegraphics[width=.112\linewidth, height=0.45in]{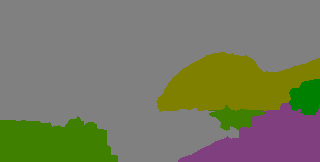} 
     \includegraphics[width=.112\linewidth, height=0.45in]{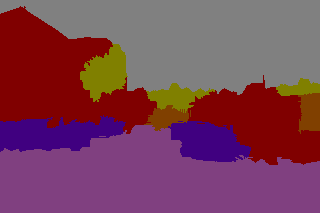}\\

     \includegraphics[width=.112\linewidth, height=0.45in]{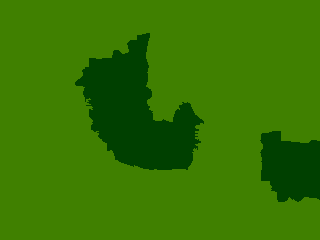}
     \includegraphics[width=.112\linewidth, height=0.45in]{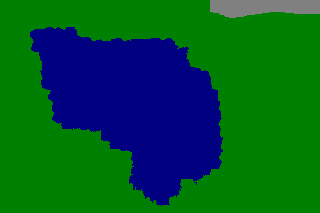}     
	\includegraphics[width=.112\linewidth, height=0.45in]{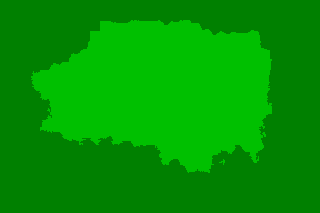}
    \includegraphics[width=.112\linewidth, height=0.45in]{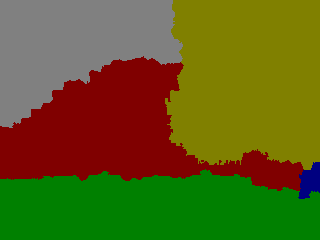}
     \includegraphics[width=.112\linewidth, height=0.45in]{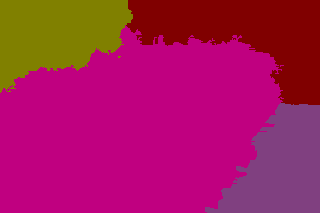}
     \includegraphics[width=.112\linewidth, height=0.45in]{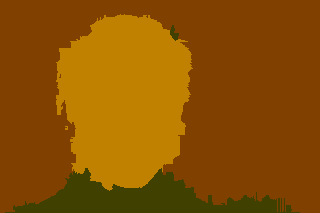}   
     \includegraphics[width=.112\linewidth, height=0.45in]{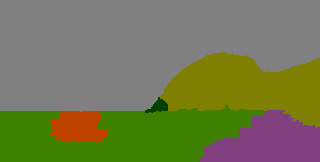} 
     \includegraphics[width=.112\linewidth, height=0.45in]{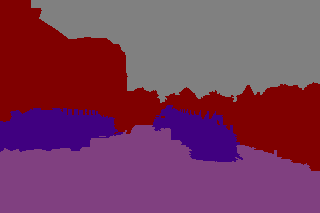}\\

\caption{Segmentation examples on MSRC. 1st row: Test images; 2nd row: Ground truth; 
3rd row: Segmentation results produced by \ssvm based \crf learning with bag-of-words feature;
4th row: Segmentation results produced by \ssvm based \crf learning with unsupervised feature learning; 
5th row: Segmentation results produced by our method with co-occurrence pairwise potentials. }
    \label{fig:seg_msrc}      
\end{figure}

\paragraph{MSRC-21}
The compared results with features of different layers are summarized in Table~\ref{tab:seg_msrc_baseline}.  
Different from the binary cases as Weizmann horse and Graz-02,  features of the 7th layer perform the best, which may results from the fact that MSRC is much more difficult due to the many categories. 
Figure~\ref{fig:seg_msrc} shows some qualitative results of 
\ssvm based \crf learning with different features, from which similar conclusions can be drawn.

\begin{table}[!t]
\centering
\resizebox{0.96\linewidth}{!}
{
\begin{tabular}{ c c| c c c c c c c c c c c c c c c c c c c c c |c c }
& &\rotatebox{90}{building}  &\rotatebox{90}{grass}  &\rotatebox{90}{tree}  &\rotatebox{90}{cow}  &\rotatebox{90}{sheep}  &\rotatebox{90}{sky}  &\rotatebox{90}{aeroplane}  &\rotatebox{90}{water}  &\rotatebox{90}{face}  &\rotatebox{90}{car}   &\rotatebox{90}{bicycle}  &\rotatebox{90}{flower} &\rotatebox{90}{sign}  &\rotatebox{90}{bird}  &\rotatebox{90}{book}  &\rotatebox{90}{chair}  &\rotatebox{90}{road}  &\rotatebox{90}{cat}  &\rotatebox{90}{dog}  &\rotatebox{90}{body}  &\rotatebox{90}{boat}  &\rotatebox{90}{\bf{Average}} &\rotatebox{90}{\bf{Global}} \\ 
\hline  \hline
\multirow{5}{*}{\svm} 
&BoW &61  &87  &60  &29  &47  &83  &56  &66  &60  &54  &66  &53  &68  &7  &61  &33  &51  &27  &35  &19  &29  &50.1  &62.7 \\
&UFL &57    &95    &77    &55    &59    &96    &56    &70    &61    &41    &67    &65    &31    &17    &67    &30    &75   &52    &26    &32     &6  &54.1  &69.5 \\
&L5  &77    &91    &86    &79    &83    &95    &80    &85    &81    &76    &84    &81    &52    &55    &82    &64    &83    &81    &63    &68   &25  &74.8  &82.1 \\
&L6  &78    &95    &88    &81    &87    &95    &83    &88    &86    &75    &86  &83    &55    &58    &86    &69    &85    &84    &67    &72    &28   &77.6   &84.9 \\
&L7  &80    &98    &89    &82    &91    &96    &86    &87    &89    &76    &86  &86    &58    &59    &87    &68    &87    &85    &67    &74    &31  &\textbf{79.0}   &\textbf{86.0}  \\
\hline
\multirow{5}{*}{\ssvm} 
&BoW  &65  &89  &87  &64  &74  &90  &58  &75  &78  &56  &85  &54  &55  &6  &60  &14  &66  &50  &35  &38  &8  &57.4  &70.7  \\
&UFL &70  &97  &87  &69  &77  &98  &45  &75  &77  &49  &86  &82  &26  &12  &81  &40  &79  &49  &14  &47  &1  &60.1  &76.1\\
&L5   &71    &97    &92    &86    &95    &98    &94    &82    &93    &80    &95  &92    &76    &65    &94    &72    &89    &87    &71    &78    &51  &83.9  &86.9 \\
&L6   &71    &94    &93    &89   &96    &96    &95    &85    &92    &85    &95  &90    &71    &68    &94    &77    &92    &93    &75    &81    &54  &85.8  &87.3 \\
&L7   &71    &95    &92    &87    &98    &97    &97    &89    &95    &85    &96    &94    &75    &76    &89    &84    &88    &97    &77    &87    &52   &\textbf{86.7} &\textbf{88.5} \\
\hline
\end{tabular}
}
\caption{Segmentation results on the MSRC-21 dataset.
We report the pixel-wise accuracy for each category as well as the average per-category scores and the global pixel-wise accuracy (\%). 
Deep learning performs significantly better than the BoW feature and the unsupervised feature learning, with \ssvm based \crf learning using features of the 7th layer of the deep CNN achieving the best results. 
\ssvm based \crf learning performs far better than \svm.
}
\label{tab:seg_msrc_baseline}
\end{table}

\subsection{State-of-the-art comparison}
\label{sec:exp2}
Based on the above evaluation, we choose the best performed 6th layer for the binary (Weizmann horse and Graz-02) and 7th layer features for the multi-class datasets (MSRC-21, Stanford Background and VOC 2011)  to learn \crf and compare with state-of-the-art results in this section. 
For the three multi-class datasets, we add the results of incorporating the mutex and co-occurrence pairwise potentials introduced in Section~\ref{sec:co-occur}.

\paragraph{Binary datasets}
Table~\ref{tab:seg_binary} shows the compared segmentation results on the Weizmann horse and the Graz-02 datasets. We use a different evaluation metric for comparison on the Graz-02 dataset, which is the F-score ($F=2pr/(p+r)$, where $p$ is the precision and $r$ is the recall) for each class and the average over classes. In both cases, our method outperforms all the compared methods. 

\begin{table}
\resizebox{0.9\linewidth}{!} {
\centering
\parbox{0.85\linewidth} {
\centering
\begin{tabular}{| l | c |  c | }
\hline
 Method   &Sa  &So \\
\hline
Levin \& Weiss \cite{Levin06} &95.5 &-  \\
Cosegmentation \cite{cosegmentation10} &80.1 &- \\
Bertelli \etal \ \cite{Bertelli11}  & 94.6 &80.1 \\
Kuttel \etal \ \cite{Kuettel12} &94.7 &- \\
\hline
Ours &\textbf{95.7}	 &\textbf{84.0}	\\
\hline
\end{tabular}
}
\parbox{0.65\linewidth} {
\centering
\begin{tabular}{| l | c |  c | c  | c |}
\hline
 Method   &bike  &car  &people  &average \\
\hline
Marszalek \& Schimid \cite{Marszalek07}  &61.8 &53.8 &44.1 & 53.2\\
Fulkerson \etal \ \cite{Fulkerson08} &66.4  &54.7 &51.4 &57.5 \\
Aldavert \etal \ \cite{Aldavert10} &71.9 &62.9 &58.6 &64.5  \\
Kuettel \etal \ \cite{Kuettel12} &63.2  &74.8 &66.4 &68.1\\
\hline
Ours	 &\textbf{84.5}	 &\textbf{85.4}	&\textbf{80.4}  &\textbf{83.4}  \\
\hline
\end{tabular}
}
}
\caption{State-of-the-art comparison of segmentation performance (\%) on the Weizmann horse (left) and Graz-02 (right) datasets. 
}
\label{tab:seg_binary}
\end{table}

\paragraph{Multi-class datasets}
The compared global and average per-category pixel accuracies on the MSRC-21 and the Stanford Background datasets are summarized in Table~\ref{tab:seg_msrc-stanford}. 
On the MSRC dataset, our method outperforms all the methods except \cite{Roy14}. When incorporated with mutex or co-occurrence pairwise potentials in inference, we obtain further improvements. 
As expected, the co-occurrence potentials outperform the mutex potentials.
\cite{Roy14} performs slightly better than ours in terms of global accuracy (they did not report average per-category accuracy), which may results from the fact that they use a fully connected CRF while ours are not.

\begin{figure} [!t]
\centering
	\includegraphics[width=.112\linewidth, height=0.45in]{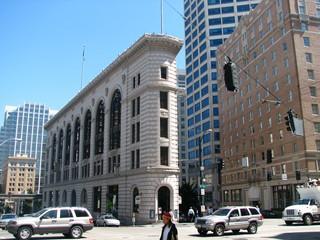}
     \includegraphics[width=.112\linewidth, height=0.45in]{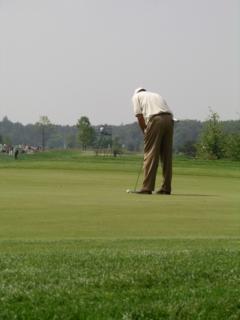}     
	\includegraphics[width=.112\linewidth, height=0.45in]{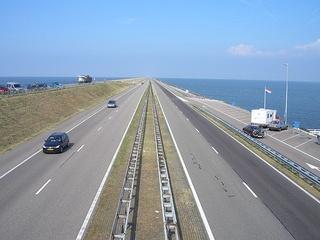}
    \includegraphics[width=.112\linewidth, height=0.45in]{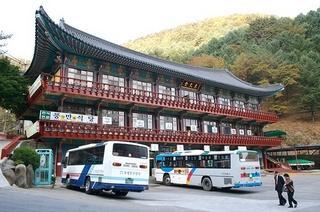}
     \includegraphics[width=.112\linewidth, height=0.45in]{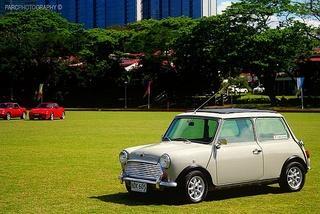}
     \includegraphics[width=.112\linewidth, height=0.45in]{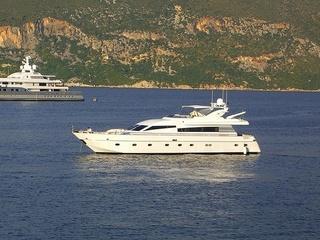}   
     \includegraphics[width=.112\linewidth, height=0.45in]{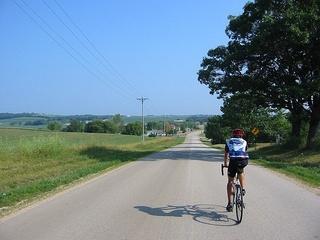} 
     \includegraphics[width=.112\linewidth, height=0.45in]{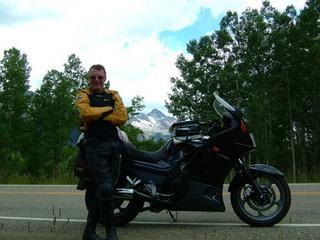}\\
     
	\includegraphics[width=.112\linewidth, height=0.45in]{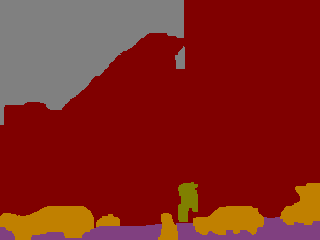}
     \includegraphics[width=.112\linewidth, height=0.45in]{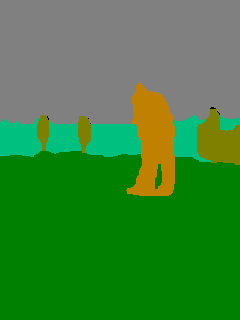}     
	\includegraphics[width=.112\linewidth, height=0.45in]{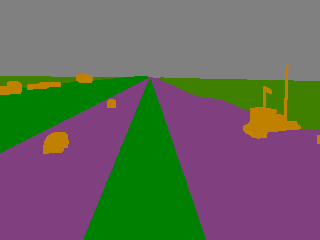}
	\includegraphics[width=.112\linewidth, height=0.45in]{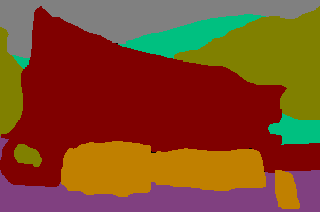}
     \includegraphics[width=.112\linewidth, height=0.45in]{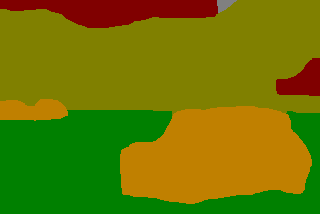}
     \includegraphics[width=.112\linewidth, height=0.45in]{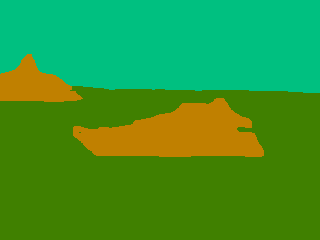}   
     \includegraphics[width=.112\linewidth, height=0.45in]{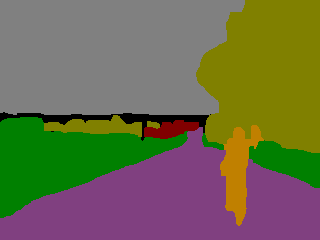} 
     \includegraphics[width=.112\linewidth, height=0.45in]{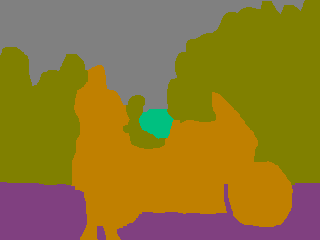}\\

     \includegraphics[width=.112\linewidth, height=0.45in]{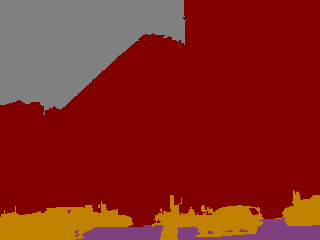}
     \includegraphics[width=.112\linewidth, height=0.45in]{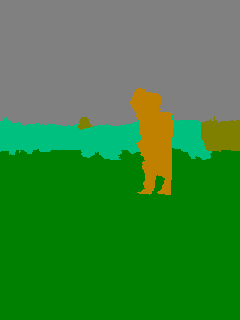}     
	\includegraphics[width=.112\linewidth, height=0.45in]{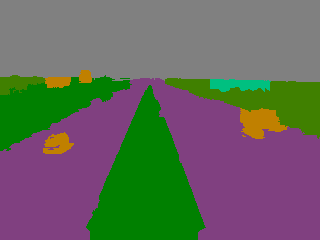}
    \includegraphics[width=.112\linewidth, height=0.45in]{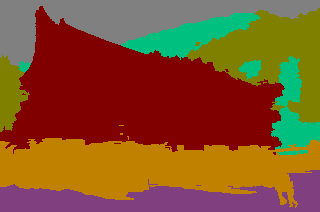}
     \includegraphics[width=.112\linewidth, height=0.45in]{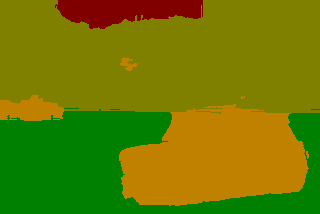}
     \includegraphics[width=.112\linewidth, height=0.45in]{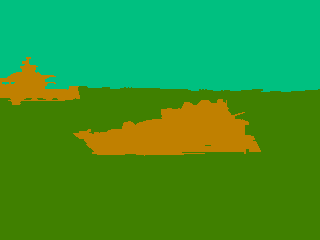}   
     \includegraphics[width=.112\linewidth, height=0.45in]{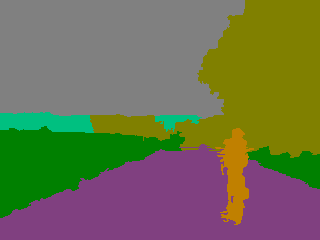} 
     \includegraphics[width=.112\linewidth, height=0.45in]{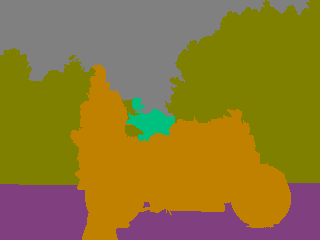}\\

     \includegraphics[width=.37\linewidth, height=0.18in]{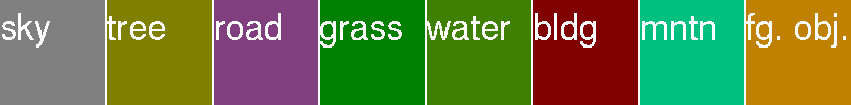}

\caption{Segmentation examples on the Stanford Background dataset. 1st row: Test images; 2nd row: Ground truth; 
3rd row: Segmentation results produced by our method with co-occurrence pairwise potentials. }
    \label{fig:seg_stanford}      
\end{figure}

As for the Stanford Background dataset, we can see that our method performs better than \cite{Lecun13} and outperforming all the others.
The work of \cite{Lecun13} trains a 3-stage multiscale convolutional network on the training images while we directly transfer the deep CNN trained on the ImageNet to here sparing the effort of network training.
Adding mutex potentials to our method do not bring any performance boost. By further investigations, we found that this is because there is only eight categories (one of which is the ambiguous foreground category) in this dataset, which leads to the fact that the only mutex information obtained is that grass, water and road can not appear above sky.  
Instead, our co-occurrence potentials perform much better, leading to further performance boost.
We show some segmentation examples in Figure~\ref{fig:seg_stanford}.

\begin{figure} [!t]
\centering
	\includegraphics[width=.112\linewidth, height=0.45in]{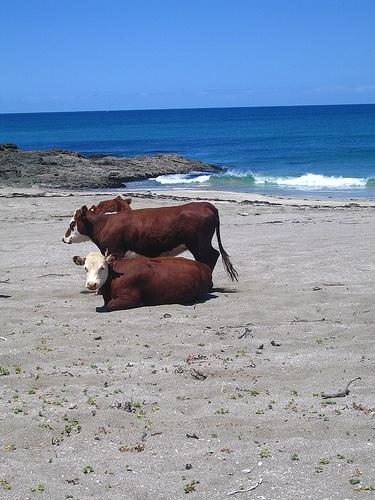}
     \includegraphics[width=.112\linewidth, height=0.45in]{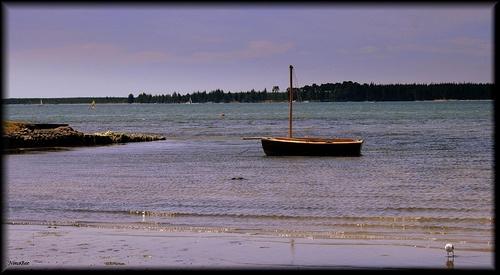}     
	\includegraphics[width=.112\linewidth, height=0.45in]{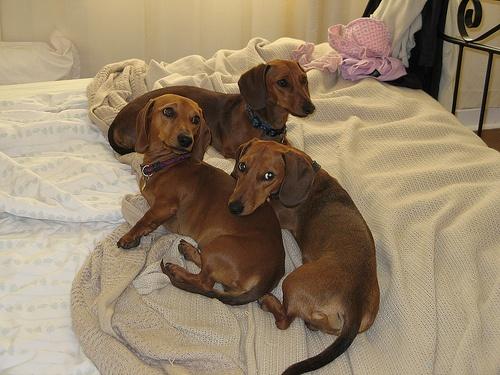}
    \includegraphics[width=.112\linewidth, height=0.45in]{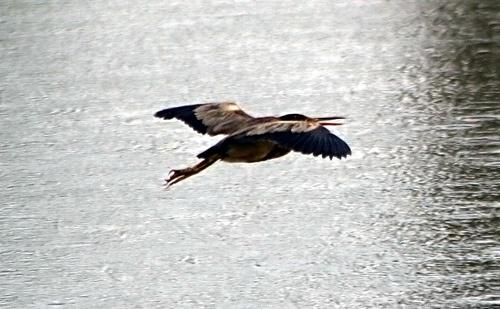}
     \includegraphics[width=.112\linewidth, height=0.45in]{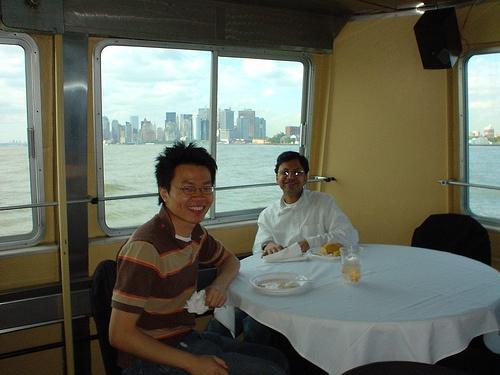}
     \includegraphics[width=.112\linewidth, height=0.45in]{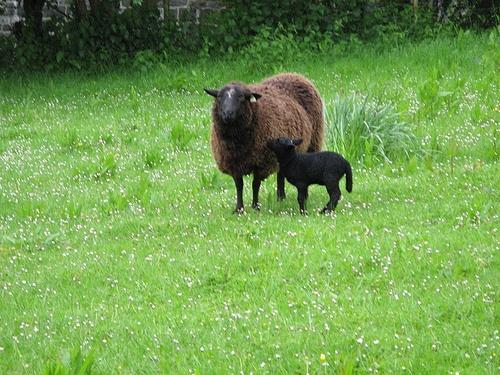}   
     \includegraphics[width=.112\linewidth, height=0.45in]{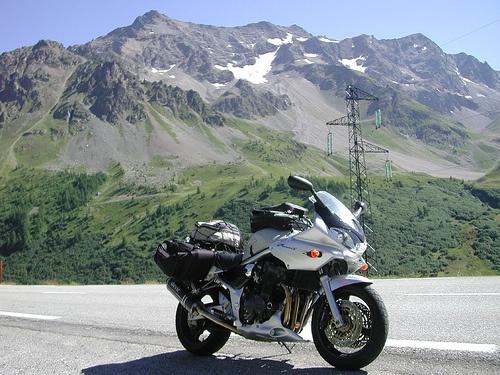} 
     \includegraphics[width=.112\linewidth, height=0.45in]{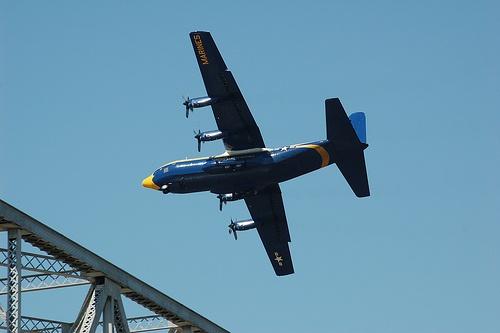}\\
     
	\includegraphics[width=.112\linewidth, height=0.45in]{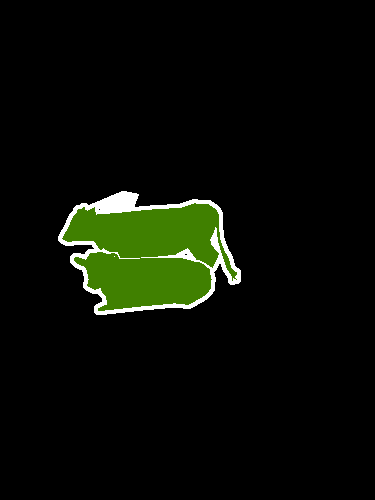}
     \includegraphics[width=.112\linewidth, height=0.45in]{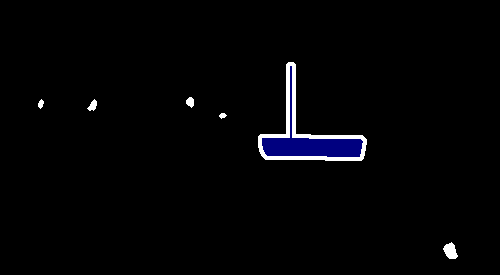}     
	\includegraphics[width=.112\linewidth, height=0.45in]{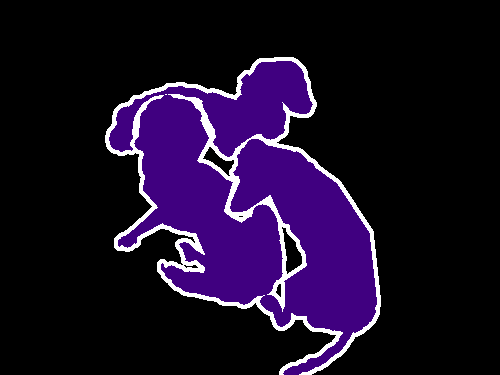}
	\includegraphics[width=.112\linewidth, height=0.45in]{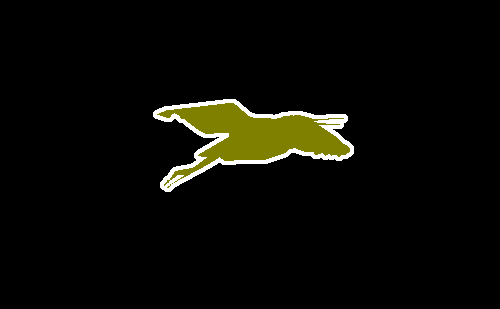}
     \includegraphics[width=.112\linewidth, height=0.45in]{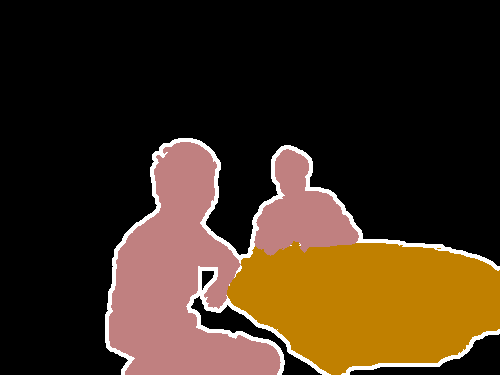}
     \includegraphics[width=.112\linewidth, height=0.45in]{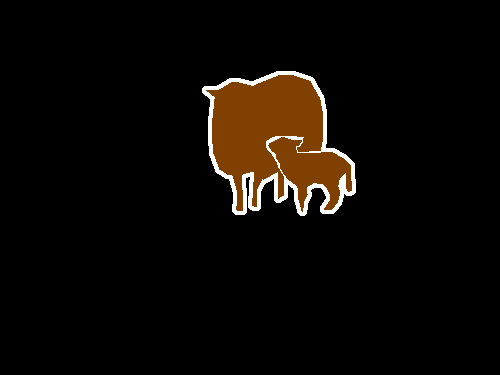}   
     \includegraphics[width=.112\linewidth, height=0.45in]{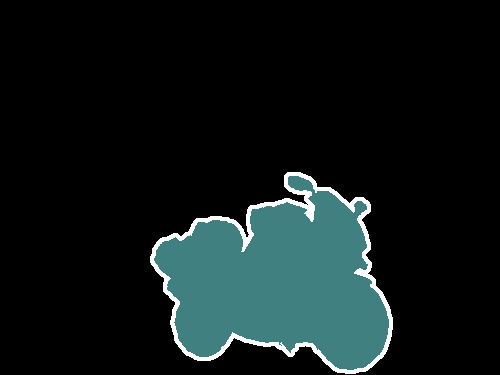} 
     \includegraphics[width=.112\linewidth, height=0.45in]{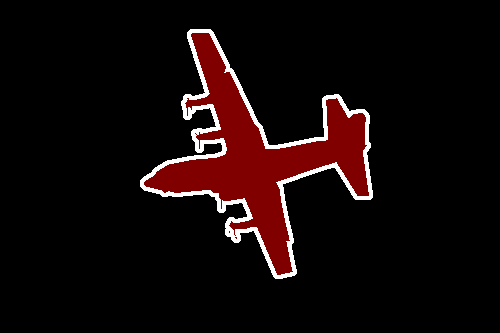}\\

     \includegraphics[width=.112\linewidth, height=0.45in]{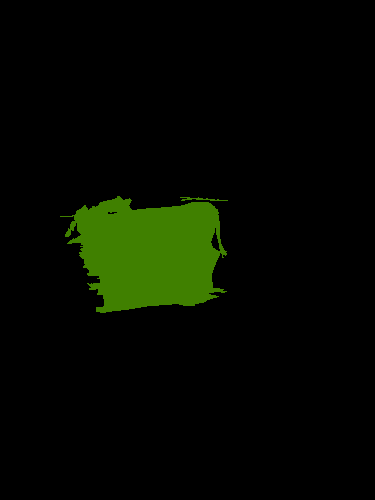}
     \includegraphics[width=.112\linewidth, height=0.45in]{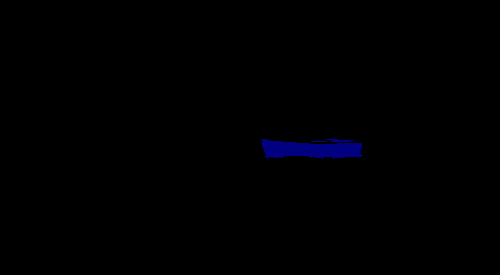}     
	\includegraphics[width=.112\linewidth, height=0.45in]{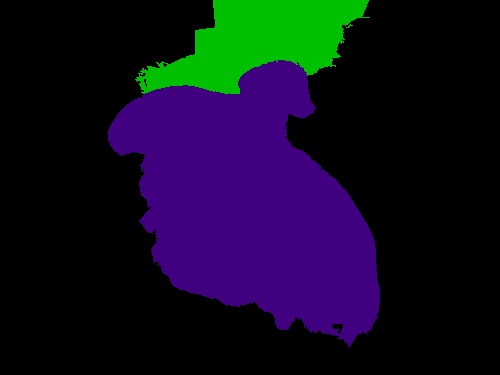}
    \includegraphics[width=.112\linewidth, height=0.45in]{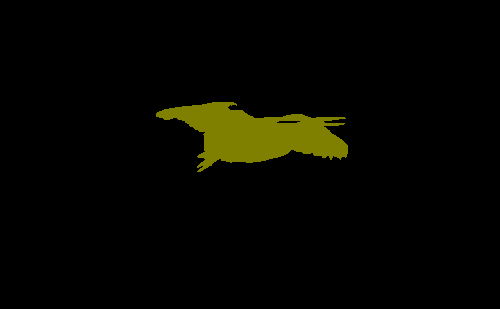}
     \includegraphics[width=.112\linewidth, height=0.45in]{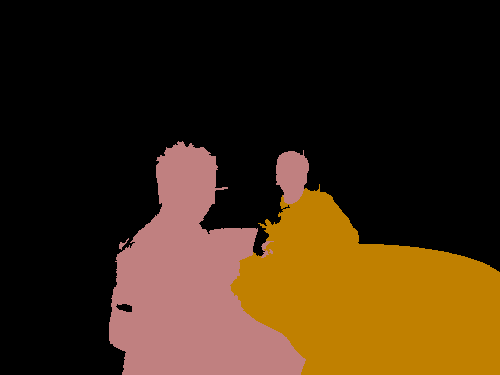}
     \includegraphics[width=.112\linewidth, height=0.45in]{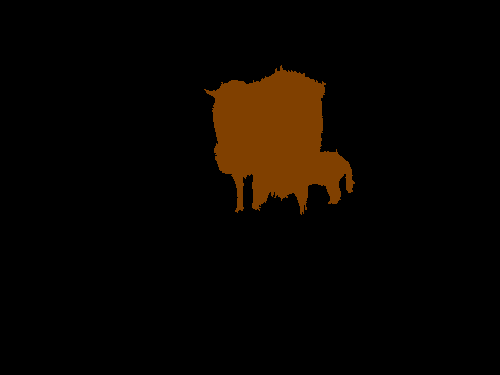}   
     \includegraphics[width=.112\linewidth, height=0.45in]{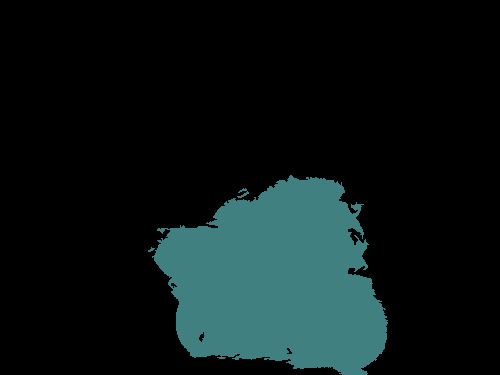} 
     \includegraphics[width=.112\linewidth, height=0.45in]{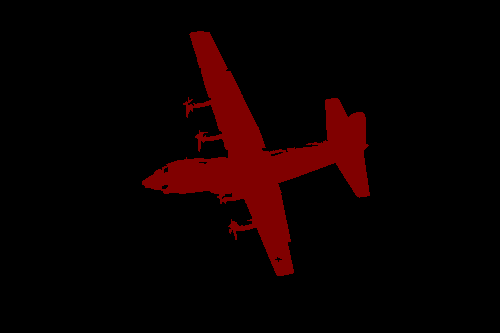}\\

\caption{Segmentation examples on the VOC 2011 dataset. 1st row: Test images; 2nd row: Ground truth; 
3rd row: Segmentation results produced by our method with co-occurrence pairwise potentials. }
    \label{fig:seg_voc11}      
\end{figure}

\begin{figure} [!t]
\centering
	\includegraphics[width=.112\linewidth, height=0.45in]{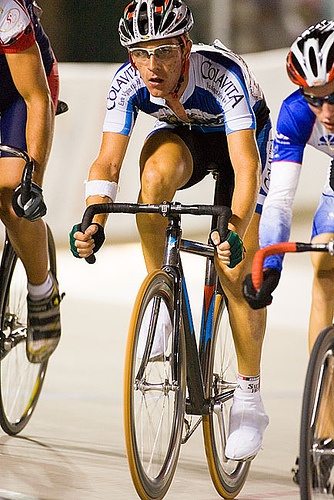}
     \includegraphics[width=.112\linewidth, height=0.45in]{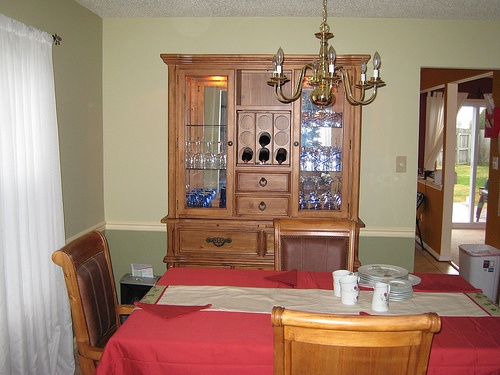}     
	\includegraphics[width=.112\linewidth, height=0.45in]{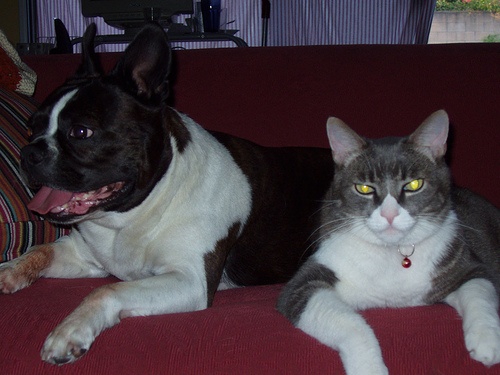}
     \includegraphics[width=.112\linewidth, height=0.45in]{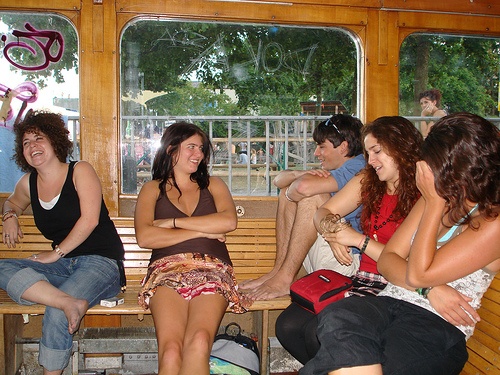}
     \includegraphics[width=.112\linewidth, height=0.45in]{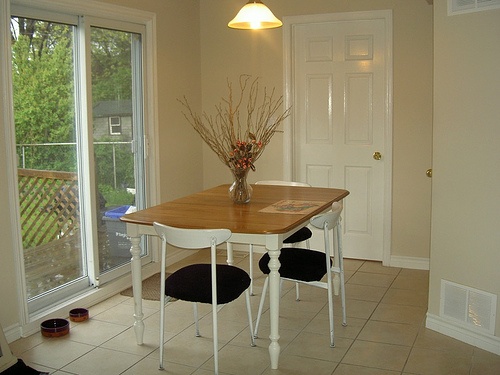}   
     \includegraphics[width=.112\linewidth, height=0.45in]{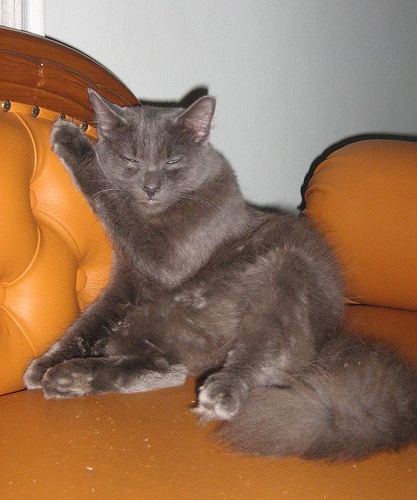}
     \includegraphics[width=.112\linewidth, height=0.45in]{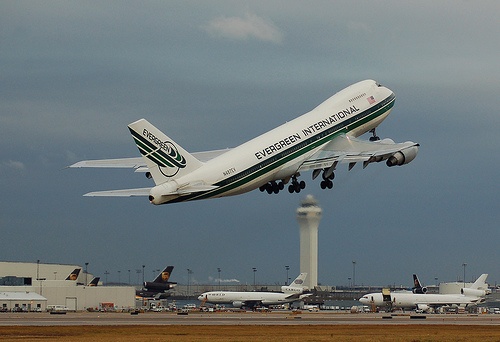} 
     \includegraphics[width=.112\linewidth, height=0.45in]{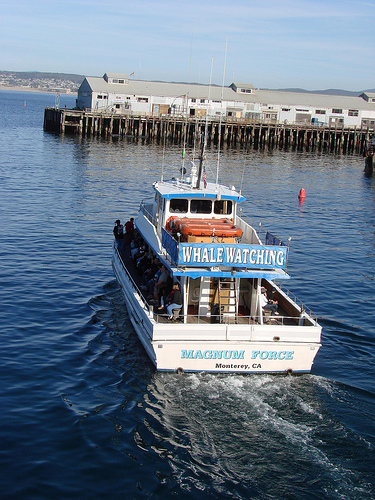}\\
     
	\includegraphics[width=.112\linewidth, height=0.45in]{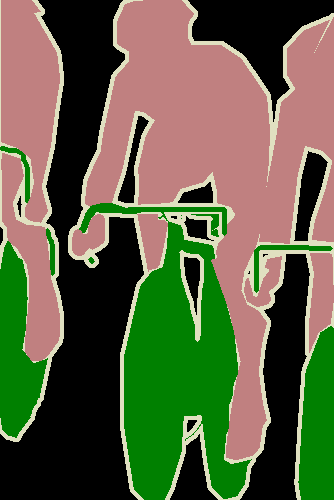}
     \includegraphics[width=.112\linewidth, height=0.45in]{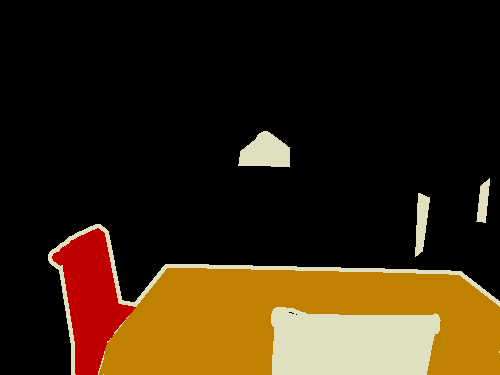}     
	\includegraphics[width=.112\linewidth, height=0.45in]{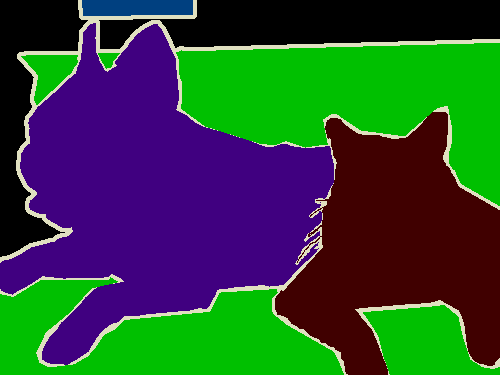}
     \includegraphics[width=.112\linewidth, height=0.45in]{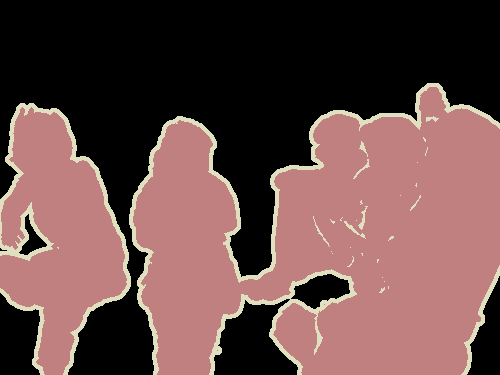}
     \includegraphics[width=.112\linewidth, height=0.45in]{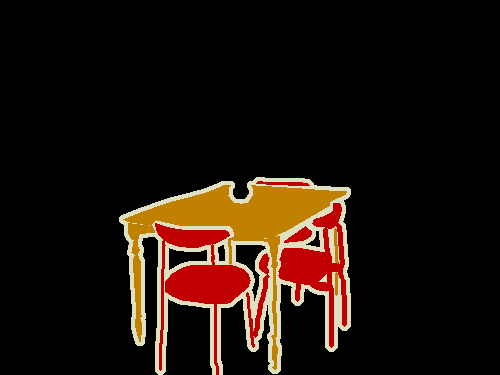} 
     \includegraphics[width=.112\linewidth, height=0.45in]{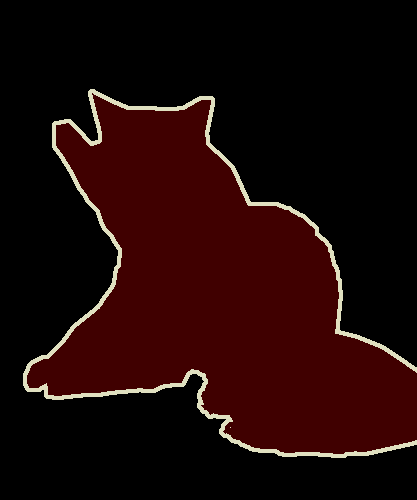}  
     \includegraphics[width=.112\linewidth, height=0.45in]{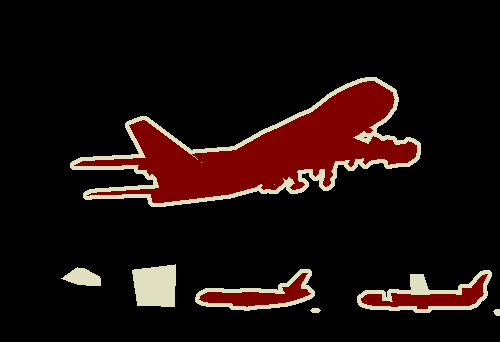} 
     \includegraphics[width=.112\linewidth, height=0.45in]{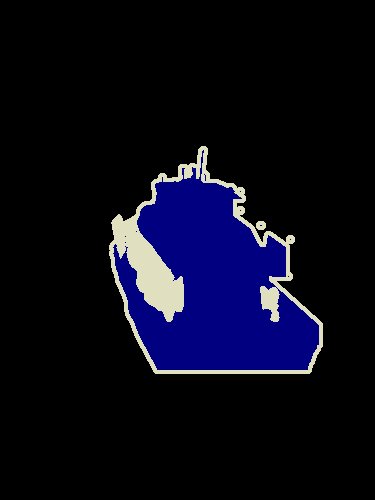}\\

     \includegraphics[width=.112\linewidth, height=0.45in]{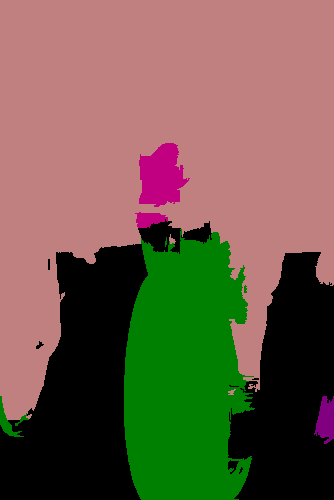}
     \includegraphics[width=.112\linewidth, height=0.45in]{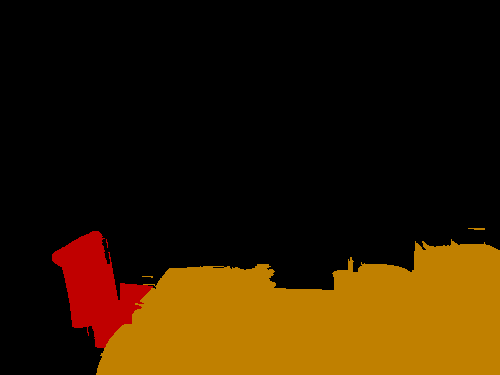}     
	\includegraphics[width=.112\linewidth, height=0.45in]{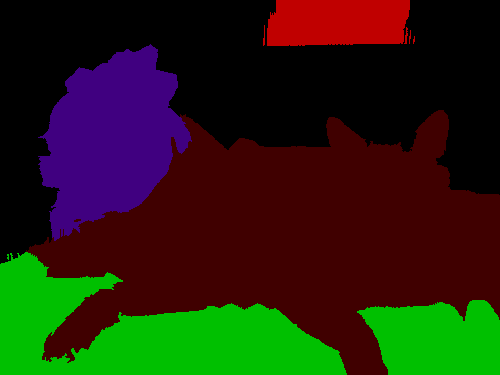}
     \includegraphics[width=.112\linewidth, height=0.45in]{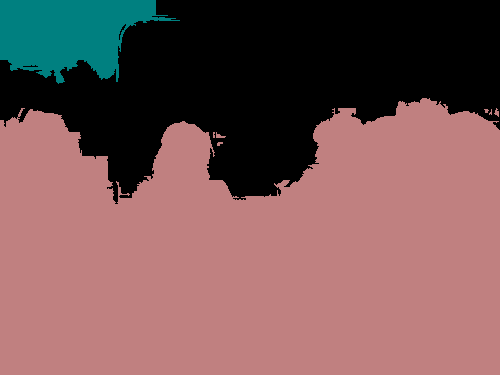}
     \includegraphics[width=.112\linewidth, height=0.45in]{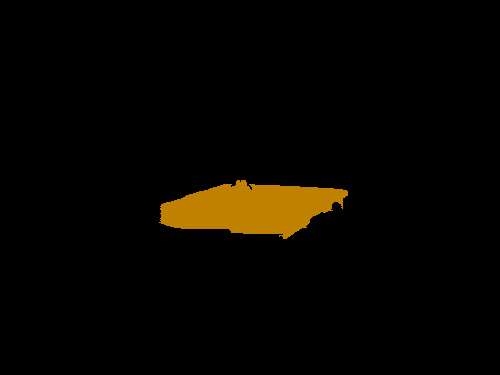} 
     \includegraphics[width=.112\linewidth, height=0.45in]{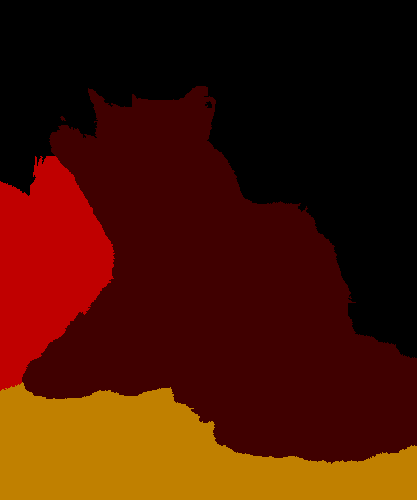}  
     \includegraphics[width=.112\linewidth, height=0.45in]{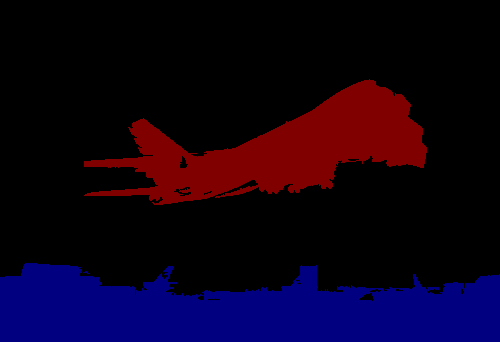} 
     \includegraphics[width=.112\linewidth, height=0.45in]{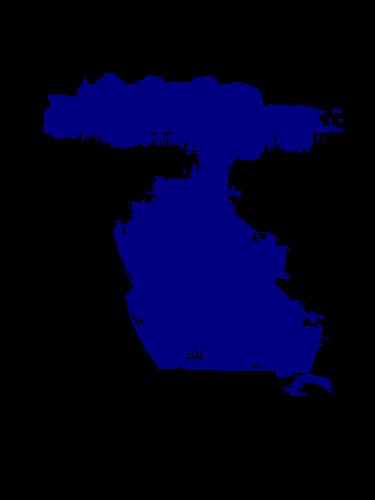}\\

\caption{Failure examples on the VOC 2011 dataset. 1st row: Test images; 2nd row: Ground truth; 
3rd row: Segmentation results produced by our method with co-occurrence pairwise potentials. }
    \label{fig:seg_voc11_fail}      
\end{figure}

The segmentation results on the PASCAL VOC 2011 validation dataset are reported in Table~\ref{tab:voc11}. 
In \cite{Ross14}, Girshick \etal \ achieved an average accuracy of 47.9 by using augmented training data and extra annotation set.
Here we did not use any extra dataset but only the VOC training set.
By introducing mutex or co-occurrence pairwise potentials, constant improvements are observed on most of the categories.
As expected, our co-occurrence potential again outperforms the mutex potential.
In Table \ref{tab:voc11_comp}, we compare with the recent work of Carreira \etal \ \cite{Carreira14}, which performed evaluations with the same settings as ours (using the train/val set).
Our method achieves the same accuracy as  \cite{Carreira14}.
Note that the dimension of the feature descriptors used in \cite{Carreira14} is tens of thousands of ($33589$) while ours is $4096$.
Qualitative examples and some failure cases  are shown in Figure~\ref{fig:seg_voc11} and  Figure~\ref{fig:seg_voc11_fail}.

\begin{table}[t]
\resizebox{0.9\linewidth}{!}{
\centering
\parbox{0.8\linewidth}{
\centering
\begin{tabular}{| l |c |c | }
\hline
Method &Global (\%) &Average (\%) \\
\hline  
Shotton \etal \ \cite{Shotton08} &72   &67  \\
Ladicky \etal \ \cite{Ladicky09}  &86   &75  \\
Munoz \etal \ \cite{Munoz10}   &78  &71 \\
Gonfaus \etal \ \cite{Harmony10} &77   &75 \\
Lucchi \etal \ \cite{Lucchi12} &73  &70 \\
Yao \etal \ \cite{Yao12}  &86.2  &79.3   \\
Lucchi \etal \ \cite{Lucchi13}   &83.7   &78.9  \\
Ladicky \etal \ \cite{Ladicky13}  &87  &77  \\
Roy \etal \ \cite{Roy14}  &\textbf{91.5}  &-  \\ 
\hline
Ours  &88.5  &86.7   \\
Ours (mutex) &90.3  &89.2  \\
Ours (co-occur) &91.1  &\textbf{90.5} \\
\hline
\end{tabular}
}
\parbox{0.7\linewidth} {
\centering
\begin{tabular}{| l | c |  c |}
\hline
Method &Global (\%) &Average (\%) \\
\hline
Gould \etal \ \cite{Gould09} &76.4 &- \\
Munoz \etal \ \cite{Munoz10}  &  76.9  &66.2 \\
Lempitsky \etal \ \cite{Lempitsky11} &81.9 &72.4 \\
Farabet \etal \ \cite{Lecun13} &81.4 &76.0 \\
Roy \etal \ \cite{Roy14} &81.1 &- \\
\hline
Ours &82.6  &76.2 \\
Ours (mutex) &82.6 &76.3 \\
Ours (co-occur) &\textbf{83.5} &\textbf{76.9}  \\
\hline
\end{tabular}
}
}
\caption{State-of-the-art comparison of global and average per-category pixel accuracy on the MSRC-21 (left) and the Stanford Background (right) datasets.
}
\label{tab:seg_msrc-stanford}
\end{table}

\begin{table}
\centering
\resizebox{1\linewidth}{!} {
\centering
\begin{tabular}{| l | c| c c c c c c c c c c c c c c c c c c c c  |  c |}
\hline
\textbf{VOC 2011 val}  &bg &aero &bike &bird &boat &bottle &bus &car &cat &chair &cow &table &dog &house &mbike &person &plant &sheep &sofa &train &tv &mean\\
\hline
Ours  &78.3  &43.9  &20.4  &23.2  &22.7  &24.6  &42.2  &41.0  &36.1  &12.6  &24.9  &19.8  &25.0  &23.8  &38.6  &53.3  &20.0  &36.6  &20.2  &38.1  &24.6  &31.9\\
Ours (mutex) &79.8  &53.1  &\textbf{23.8}  &\textbf{26.4}  &\textbf{28.8}  &\textbf{28.6}  &51.6  &48.2  &\textbf{37.8}  &13.1  &29.7  &22.3  &\textbf{28.4}  &29.6  &45.2  &52.7  &21.0  &46.2  &20.9  &46.2  &29.6  &36.3  \\ 
Ours (co-occur) &\textbf{81.5}  &\textbf{55.7}  &23.6  &24.0  &27.7  &27.3  &\textbf{52.8}  &\textbf{54.1}  &37.1  &\textbf{14.9}  &\textbf{37.1}  &\textbf{28.6}  &22.9  &\textbf{33.1}  &\textbf{49.7}  &\textbf{54.2}  &\textbf{27.4}  &\textbf{49.3}  &\textbf{22.3}  &\textbf{49.3}  &\textbf{30.9}  &\textbf{38.3} \\
\hline
\end{tabular}
}
\caption{Results of per-category and mean segmentation accuracy (\%) on the PASCAL VOC 2011 validation dataset. Best results are bold faced.}
\label{tab:voc11}
\end{table}

\begin{table}[t]
\centering
\resizebox{0.3\linewidth}{!}{
\begin{tabular}{| l |c | }
\hline
Method  &Mean (\%) \\
\hline  
HOG \cite{Carreira14}   &14.1  \\
SIFT-PCA-FISHER \cite{Carreira14}   &31.9  \\
O$_2$P \cite{Carreira14}   &\textbf{38.3}\\
\hline
Ours (co-occur) &\textbf{38.3} \\
\hline
\end{tabular}
}
\caption{Comparison of the mean segmentation accuracy (\%) on the PASCAL VOC 2011 validation dataset.}
\label{tab:voc11_comp}
\end{table}

\section{Conclusion}
We propose to learn \crf using \ssvm based on features learned from a pre-trained deep convolutional neural network for image segmentation.
The deep CNN is trained on ImageNet and proved to perform exceptionally well when transferred to object segmentation.
We learn the \crf in the large margin framework by \ssvm,
and then conduct inference with co-occurrence pairwise potentials incorporated.
Extensive experimental evaluations on the Weizmann horse, Graz-02, MSRC-21, Stanford Background and the PASCAL VOC 2011 dataset demonstrate the advantages of our method and provide new baselines for further research.

\section*{References}

\bibliographystyle{ieee}

\bibliography{seg}

\end{document}